\documentclass[11pt,letterpaper]{article}

\usepackage[margin=1.1in]{geometry}
\usepackage[utf8]{inputenc}
\usepackage[T1]{fontenc}
\usepackage{lmodern}
\usepackage{textcomp}
\usepackage{microtype}

\DeclareUnicodeCharacter{2212}{\textminus}

\usepackage{amsmath,amssymb,amsfonts,bm}

\usepackage{amsmath,amsfonts,bm}

\def\eqref#1{equation~\ref{#1}}

\def\1{\bm{1}}

\DeclareMathAlphabet{\mathsfit}{\encodingdefault}{\sfdefault}{m}{sl}
\SetMathAlphabet{\mathsfit}{bold}{\encodingdefault}{\sfdefault}{bx}{n}

\usepackage[dvipsnames,svgnames,table]{xcolor}

\definecolor{pastgreen}{HTML}{C7EDCC}
\definecolor{pastyellow}{HTML}{FFF3CD}
\definecolor{pastred}{HTML}{F8D7DA}

\usepackage{graphicx}
\usepackage{booktabs}
\usepackage{wrapfig}
\usepackage{nicefrac}
\usepackage{pifont}
\usepackage{placeins}

\usepackage{algorithm}
\usepackage{algpseudocode}
\usepackage{listings}
\usepackage{fvextra}

\usepackage{subcaption}

\fvinlineset{breaklines=true,breakanywhere=true}

\usepackage{epigraph}

\usepackage{authblk}

\usepackage[authoryear,round]{natbib}
\usepackage{hyperref}

\usepackage{enumitem}

\hypersetup{
  pdftitle={ArtifactArena: Evaluating Models by What They Build in the Physical World},
  pdfauthor={Kushagra Tiwary, David Mayo, Nikhil Behari, Xiangzhou Sun, Abdulrahman Alabdulkareem, Isaac Galatzer-Levy, Boris Katz, Brian Cheung},
  pdfsubject={Research preprint} 
}

\title{%
  \LARGE\bfseries
  ArtifactArena: Evaluating Models by\\
  What They Build in the Physical World
}
\author[$\ast$,1,2]{%
  \href{https://www.kushagratiwary.com/}{Kushagra Tiwary}
}
\author[$\ast$,1]{%
  \href{http://david-mayo.com/}{David Mayo}
}
\author[1,2]{%
  \href{https://nikhilbehari.github.io/}{Nikhil Behari}
}
\author[1]{Xiangzhou Sun}
\author[1]{%
  \href{https://arkareem.com/}{Abdulrahman Alabdulkareem}
}
\author[3]{%
  \href{https://med.nyu.edu/faculty/isaac-r-galatzer-levy}{Isaac Galatzer-Levy}
}
\author[1]{%
  \href{https://people.csail.mit.edu/boris/boris.html}{Boris Katz}
}
\author[$\dagger$,1,4]{%
  \href{https://briancheung.github.io/}{Brian Cheung}
}

\affil[1]{\href{https://www.csail.mit.edu/}{MIT CSAIL}}
\affil[2]{\href{https://www.media.mit.edu/groups/camera-culture/overview/}{Camera Culture}, MIT Media Lab}
\affil[3]{\href{https://med.nyu.edu/departments-institutes/psychiatry/}{Department of Psychiatry}, NYU Grossman School of Medicine}
\affil[4]{\href{https://discolab.org/}{Discovery Lab}, UCSF}
\affil[$\ast$]{Equal contribution}
\affil[$\dagger$]{Corresponding PI. Correspondence:
  \href{mailto:ktiwary@mit.edu,dmayo2@mit.edu,bcheung@ucsf.edu}
     {\texttt{ktiwary@mit.edu, dmayo2@mit.edu, bcheung@ucsf.edu}}
}

\affil[ ]{Code: 
  \href{https://github.com/orgs/ArtifactArena/repositories}{\texttt{GitHub}}
}
\date{Last updated: October 5th, 2026}
\newif\ifshowcomments
\showcommentsfalse
\begin{document}

\raggedbottom
\maketitle

\begin{abstract}
To evaluate the frontier, we must measure models not by what they say,
but by what they can engineer and build in grounded physical
environments. We introduce \textsc{ArtifactArena}, an open-ended
platform where models face a physically grounded hardware-software
co-design challenge: engineering fully functional robots to compete
in a simulated arena. We evaluate a frontier model's zero-shot,
verifier guided refinement, and open-ended physical design
capabilities through three harnesses that refine their bot artifacts
based on text descriptions, physics simulator feedback, and gameplay
data. We benchmark these capabilities with an Elo ranking of frontier
models derived from head-to-head tournaments between their artifacts.
By releasing this framework and tournament infrastructure for ongoing
community submissions, we establish a living, non-saturating testbed
to continuously measure the expanding limits of open-ended
intelligence in the physical world.
\begin{center}
  \small\url{https://artifactarena.ai}
\end{center}
\end{abstract}
\vspace{-3mm}
\epigraph
  {``Well done is better than well said.''}
  {---Benjamin Franklin}

\section{Introduction}

In 2002, a New Caledonian crow named Betty was shown a small bucket of food at the bottom of a vertical tube and a straight piece of wire. She picked up the wire, wedged one end against the tube, bent it into a hook, and fished the bucket out \citep{weir2002shaping}. What struck the researchers watching was not anything Betty said about her solution (Betty said nothing). It was the physical artifact, the hook, that evinced Betty’s intelligence.

Frontier model evaluations lag model capabilities: popular benchmarks such as MMLU~\citep{hendrycks2021mmlu}, MMLU-Pro \citep{wang2024mmlu}, GPQA Diamond~\citep{rein2024gpqa} or code-generation benchmarks such as HumanEval~\citep{chen2021humaneval} test models against static question sets and answer keys which are now effectively saturated. To alleviate this issue, preference arenas such as LMArena~\citep{chiang2024chatbotarena} and DesignArena replace fixed answer keys with head-to-head (h2h) competition between models judged by humans or other models. The lack of an objective verifier often leads to performative behavior \citep{3666122.3668142}. As a result, functional evaluations such as SWE-Bench~\citep{jimenez2024swebench} and KernelBench~\citep{ouyang2025kernelbenchllmswriteefficient} test for functional correctness of model-generated code by running it against fixed tests, performance and efficiency baselines. However, they continue to evaluate models against predefined task objectives. Arena based evaluations like SWE-Arena~\citep{zhao2025swearenainteractiveplatformevaluating} and CodeClash~\citep{yang2025codeclashbenchmarkinggoalorientedsoftware} share a similar philosophy of model evaluation through h2h competition however they are confined to symbolic verifiers. 
\begin{figure}[!t]
  \centering
  \includegraphics[width=0.95\textwidth]{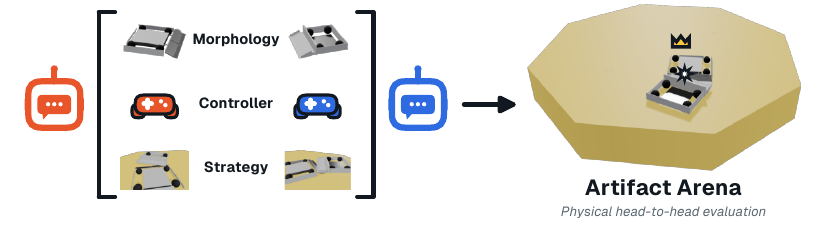}
  \caption{\textsc{ArtifactArena}: Models jointly design physically grounded artifacts: robot bodies, controllers, and strategies for head-to-head competition in MuJoCo. Artifacts win by pushing opponents off an elevated ring or pinning them under the arena's rules, grounding evaluation in the physical performance of what models create.}
  \label{fig:intro}
\end{figure}

Given these recent advances, we propose \textsc{ArtifactArena} to competitively evaluate frontier model's open-ended abilities to engineer and build verifiable artifacts that function and compete against other models' artifacts inside physically grounded simulations. In \textsc{ArtifactArena} each model is tasked with open-ended construction of a complete robot artifact consisting of a body morphology and a motion controller (Sec.~~\ref{sec:method}). While many evaluations explore physically grounded reasoning of models, \cite{chen2025physgym, chow2025physbench, bear2022physionevaluatingphysicalprediction}, we are inspired by both BattleBots \citep{battlebotsRules2022, battlebotsFandom2026, mullins2002battlebots}, and AI's success in Go~\citep{silver2016mastering} and design our arena environment based on principles that would allow creativity to emerge under physical constraints. Therefore the resulting robots are instantiated in MuJoCo~\citep{6907751} and compete in a \textsc{Last Bot Standing Game}. \textsc{Last Bot Standing} is an adversarial, objective, zero-sum game with simple rules that allow for open-ended exploration of strategies. Models are also given full creative freedom to explore a large design space of robot designs and competitive strategies, with physics as the only real constraint on exploration. Sec.~\ref{sec:results} shows how this leads to creative robot designs by models. 

The game rules are deliberately simple to allow for diverse winning strategies and objective in its outcome: a robot wins by either pushing an opponent from an elevated ring or pinning it under the arena's adjudication rules (Sec.~~\ref{sec:game_rules}). Since the model must build this robot, success requires jointly reasoning about geometry, materials, mass distribution, actuation, and feedback control, as well as competitive strategy. The outcome is determined by the artifacts’ physical interaction, rather than by a human preference label or an LLM judge. 

\textsc{ArtifactArena} deliberately keeps the physical environment and game rules fixed across evaluations, while the adversary population changes as new artifacts enter the arena. Each artifact must therefore compete against independently designed robots with different morphologies, controllers, and strategies. Holding the environment fixed makes the simulator a common objective verifier, allowing differences in performance to reflect artifact design and competitive strategy rather than changes in the task itself. As stronger artifacts enter the arena, they become new opponents that raise the bar for future submissions, allowing our arena to remain non-saturating and keep up with model capabilities. We ask models to build the strongest robots they believe can win the game and rank the models based on their artifacts’ performance. 

\begin{figure}[!tbp]
    \centering
    \includegraphics[width=\linewidth]{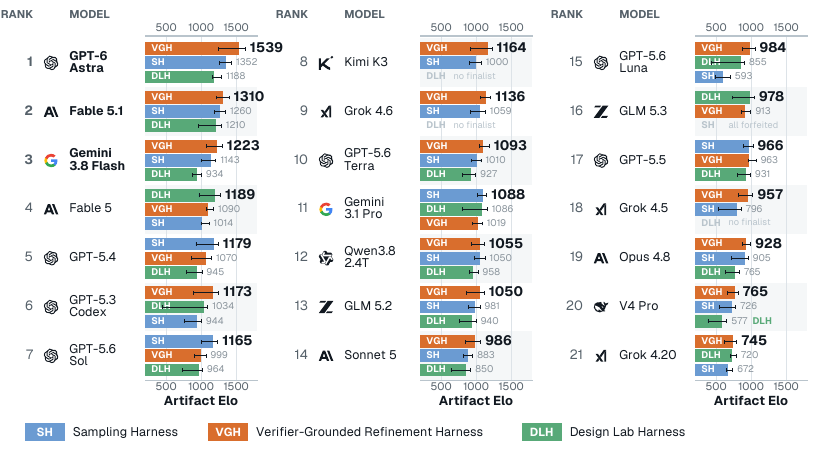}
    \caption{\textbf{\textsc{ArtifactArena} Leaderboard.} Our results show a remarkable ability in frontier models to co-design robot morphologies and controllers within physically grounded environments and constraints. We test models' engineering capabilities across three different harnesses (Sec.~\ref{sec:method_harness}) that provide models with greater agency over the artifact design process. We find that models possess a remarkable ability to zero-shot complex robot morphologies and controllers which is further enhanced by verifier grounded feedback. While our arena tests open-ended design and engineering capabilities of models, we find that open-ended agentic capabilities still lag behind verifier grounded refinement loops. Please refer to Fig.~\ref{fig:visualize_bots} to visualize the bot morphologies.}
    \label{fig:main_leaderboard}
\end{figure}

\textbf{Contributions:} (1)~\textbf{A physically grounded competitive evaluation platform.} We introduce \textsc{ArtifactArena}, an open-ended platform where models face a physically grounded hardware-software co-design challenge: engineering fully functional robots to compete in an objective zero-sum arena designed using the MuJoCo simulation environment. We maintain a live public artifact leaderboard open to community bot submissions worldwide. Our arena is harness agnostic and only the model-artifact matters: we accept both artifacts designed entirely by models and artifacts produced through human–AI collaboration. See Fig.~\ref{fig:method2}.

(2)~\textbf{A model leaderboard and systematic comparison of artifact-generation strategies.} We benchmark frontier models build capabilities using three test-time generation strategies implemented as harnesses: Sampling Harness\textsc{ (SH)}, Verifier Grounded Refinement Harness\textsc{ (VGH)}, and open-ended agentic search using a Design Lab Harness \textsc{(DLH)}. We compute-match in terms of API calls on these harnesses to study which type of harness produces the strongest artifacts. Our leaderboard shows the limits of these models frontier capabilities and should be interpreted as measuring competitive physical design and control in simulation across frontier models rather than direct real-world robotic capability. See Fig.~\ref{fig:main_leaderboard} \& Fig.~\ref{fig:example-refinement-harness}

(3) \textbf{The unreasonable effectiveness of sampling.} Our leaderboard shows that the most recent frontier models can zero-shot co-design complex robot morphologies with custom wedges using signed distance functions (SDFs) and controllers that implement charge and defend strategies while working within the physical environment. We show that this is a remarkably new capability that models have and sample this new capability to understand the extent of  these designs. (Sec.~\ref{sec:results_creativity})

(4)~\textbf{An empirical anatomy of model creativity in artifact design.} We open-source all model artifacts through the \textbf{Bot Zoo} with complete reasoning traces and design histories. We study not only the final robots but the processes through which models create them. Due to our large robot morphology and controller design space, we observe wedge, hammer, and pusher strategies implemented as both morphologies and controllers. The strongest agents also construct probes, adversarial opponents, and engineering tools to discover physical failure modes, while open-ended search can fail when self-created tests diverge from true tournament performance. (Sec.~\ref{sec:results_what_makes_models_creative})

To ensure \textsc{ArtifactArena} serves as a durable testbed for open-ended intelligence, we release the simulator, multiple harnesses, and our tournament infrastructure to support community submissions. This framework establishes a living evaluation horizon capable of continuously tracking the limits of artificial intelligence as it approaches and surpasses human-level capability in engineering and scientific endeavors.

\section{Methods}
\label{sec:method}

\begin{figure}[!t]
    \centering
    \includegraphics[width=\linewidth]{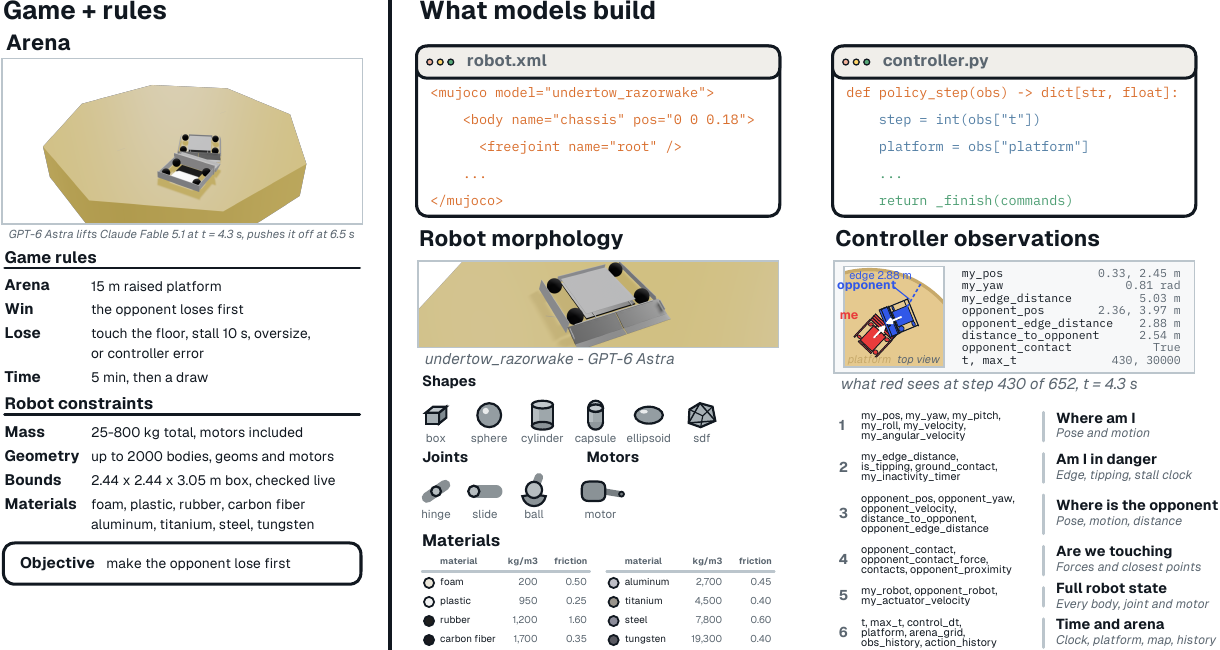}
    \caption{ArtifactArena is a functional evaluation which evaluates a model's physically grounded invention capabilities through a combat-robotics task where two autonomous robots compete on an elevated circular ring in MuJoCo with the goal of pushing the opponent off the platform. Models design complete robot artifacts, both hardware and controller. They use a set of primitives and real-world materials to design the morphology and map observations provided to the controller to motor commands towards winning the Last Bot Standing Game.}
    \label{fig:method2}
  \vspace{-5mm}
\end{figure}

The goal of \textsc{ArtifactArena} is to rank AI models based on how their generated artifacts compare against artifacts created by other models in a physically grounded arena. We hold the simulator, physical parameters, and game rules described below fixed across all evaluations so competition arises from different artifacts with diverse morphologies, controllers, and strategies. We define \textsc{ArtifactArena} as a physically grounded environment defined in MuJoCo \citep{6907751} where AI agents, denoted by \(\mathcal{A}=\{A_1,A_2,\ldots,A_n\}\), generate physically grounded robot artifacts defined by design space $X$. An artifact is a fully specified robotic system comprising a physical morphology and an autonomous control policy:
\[ x_{i}=(g_{i},c_{i},r_{i})\in X_{\mathrm{geom}}\times X_{\mathrm{ctrl}}\times X_{\mathrm{text}}=:X, \]
where \(g_{i}\) is the robot morphology, \(c_{i}\) is the autonomous controller, \(r_{i}\) is the model's design rationale. The artifact, therefore, encodes many coupled design decisions: geometry, mass distribution, joint placement, actuator limits, controller logic, feedback use, and behavior. Since model evaluation is now external to the model, our method has significantly lower evaluation costs: new models can generate artifacts that can be run against the best artifacts of older models offline (instead of re-running inference). We now discuss the arena game rules, and our methodology to rank models based on artifacts (Fig.~\ref{fig:method2}).

\subsection{The \textsc{Last Bot Standing }Game}
\label{sec:game_rules}

\textbf{Model Task.} Every model receives the same fixed prompt (Appx.~\ref{app:model_prompt}), which asks it to ``design a robot to maximize its win rate in the tournament described below'' and to return the hardware as MJCF XML and the controller as Python. We design the prompt to be descriptive rather than prescriptive: it states the game rules (Sec.~\ref{sec:game_rules}), the simulation setup in detail (to avoid any assumptions), the qualification and tournament procedure including Bradley-Terry  Elo scoring, the robot constraints and material palette (Appx.~\ref{app:material-library}), an MJCF syntax reference, and the controller observation schema (Appx.~\ref{app:harnesses}). Models are also told to think from first principles based on the MuJoCo physics world. Finally, the prompt reminds the model to optimize for expected win rate rather than simplicity, treat qualification as a constraint rather than the objective, and compare diverse morphologies and controllers before committing to one. The model must return six named sections: $r_i$ in the form of \texttt{name}, \texttt{design\_strategy}, \texttt{hardware\_plan}, \texttt{combat\_plan}, \texttt{robot\_xml} and \texttt{controller\_code} which $g_i$ and $c_i$ respectively. 

\textbf{Qualification Stage.} Once the model designs a robot artifact, it must first pass a static verifier: the XML must compile in MuJoCo, contain exactly one free joint on the root body, assign every geometry a material from the palette, stay within 25-800 kg, and fit the size box. Collision parameters, physics options, and gravity compensation in the XML are stripped to disable the model from cheating. The controller is also verified once on a live observation to check for exceptions or malformed commands. Finally the robot plays three 20s matches against a passive 342 kg plastic block that has no actuators, may translate, rotate, and topple, and is exempt from the inactivity rule. The robot must not lose in the best of three matches, or else it forfeits the tournament. Sec.~\ref{sec:method_harness} details how qualified artifacts are chosen. 

\textbf{Match.} Qualified robots are allowed to enter the tournament. A match between two robots is a head-to-head round simulated in MuJoCo \citep{6907751} on an elevated circular platform 15 m in diameter (Fig.~\ref{fig:method2}), with both robots spawned at random positions and headings. Every pair of bots plays matches with three unique seeds, once from each side, so six matches in total. Each robot has an unactuated six-degree-of-freedom root joint, allowing its body to translate and rotate freely in response to gravity, contact forces, and forces generated by its internally actuated motors driven by its controller policy code. We resolve match outcomes based on the following rules after every control step and a match that reaches 300 s without one of the robots loosing is a draw.

\textbf{Game Rules.} The \textsc{Last Bot Standing} game takes inspiration from both BattleBots \citep{battlebotsRules2022} and the ancient board game Go as both are evidence of emergent human-machine creativity in environments with simple rules and large design spaces \citep{battlebotsFandom2026, mullins2002battlebots, silver2016mastering}. A robot loses the match if any of the following conditions occur (listed in priority order):

\begin{enumerate}[nosep, topsep=2pt]
\item \emph{Controller error}: its policy raises an exception or returns malformed commands.
\item \emph{Floor contact}: any part of its body touches the floor below the platform.
\item \emph{Size violation}: its extents in its own root frame exceed 2.44 m $\times$ 2.44 m $\times$ 3.05 m.
\item \emph{Inactivity}: beginning 10 s into the match and checked every control step, the two most distant center-of-mass positions recorded over the preceding 10 s are less than 0.5 m apart.
\item \emph{Physics instability}: its body drives the simulation to non-finite forces.
\end{enumerate}


\subsection{Robot Artifact Design Space}
Frontier models must translate a high level design and competitive strategy into a robot morphology and controller that work together in MuJoCo. The morphology determines which forces the robot can apply, while the controller determines when and how to use them. We now summarize the robot design space available to the model and refer the reader to Appx.~\ref{app:robot_design_details} for the full design constraints and implementation details.


\textbf{Robot Hardware Design.} The morphology is an MJCF kinematic tree under a single free-jointed root body, with up to 2,000 bodies, geometries and actuators, and must fit within a bounding box of $2.44\,\mathrm{m} \times 2.44\,\mathrm{m} \times 3.05\,\mathrm{m}$. Bodies are built from boxes, cylinders, spheres, capsules, ellipsoids, or inline signed-distance meshes, connected by hinge, slide, or ball joints. Each geometry takes one material from the palette (Appx.~\ref{app:material-library}), which automatically sets its density (and therefore mass) based on model specified volume, and sliding friction. Motors drive internal joints, tendons, or sites, and only ever push one part of the robot against another. To allow for physically plausible motors, motor force scales with gear, and gear is charged against the mass budget, so torque is bought with weight. We also allow passive damping, friction loss, and springs on joints and tendons which let the hardware absorb impacts or store energy. 

\textbf{Controller Design.} The controller is a Python function \texttt{policy\_step(obs)} that returns a command in the range $[-1,1]$ for every actuator at each 0.01 s control step. It sees the game state through more than 50 named fields: positions, orientations, velocities, edge and opponent distances, contacts with forces and impulses, the closest geometry pairs, a rolling history of past observations and actions, and each robot's compiled bodies, joints, and motors with their current state (Fig.~\ref{fig:method2} \& Appx.~\ref{app:model_prompt}'s observation section). State persists within a match and is reset between matches, so controllers may estimate, predict, and switch modes. 

\subsection{Test-time generation strategies}
\label{sec:method_harness}

\begin{figure*}[!tbp]
  \centering
  \includegraphics[
    width=\textwidth
  ]{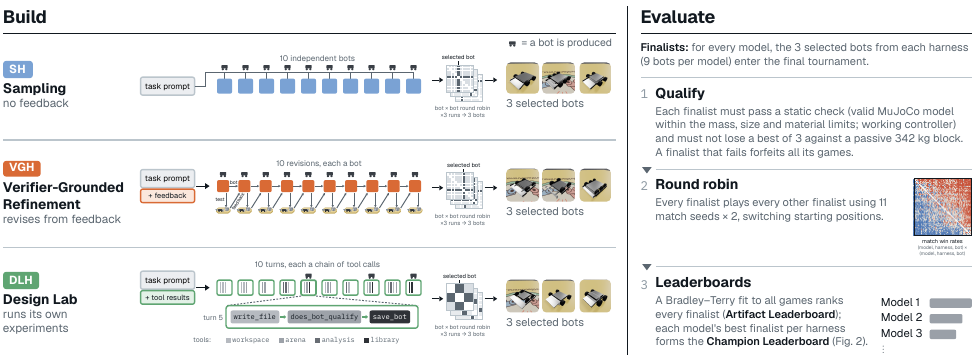}
  \caption{\textsc{ArtifactArena} evaluates models in physically grounded environments in two stages: \textsc{BUILD} phase whose goal is to output physically grounded verifiable artifacts, and \textsc{EVALUATE} phase whose goal is to rank the model through the artifact's performance. Please see Sec.~\ref{sec:method_harness} for more deatils on \textbf{Build} and Sec.~\ref{sec:methods:tournament} for more details on \textbf{Evaluate}.}
  \label{fig:functional-evaluation}
  \vspace{-5mm}
\end{figure*}

The goal of the model in \textsc{ArtifactArena} is to produce the strongest artifact it can engineer that can beat artifacts produced by other models. Therefore our arena is harness agnostic and we use this as a call to the community to build better harnesses that allow models to engineer physically grounded artifacts. We now test three test-time artifact generation strategies that differ in how much of the design loop the model controls. Figure \ref{fig:functional-evaluation} summarizes the three generation harnesses and their common evaluation procedure. See Appx.~\ref{app:harnesses} for details. For each model, we allocate 10 API calls to each harness which is run independently three times.

\textbf{Sampling Harness \textsc{(SH)}.} The model receives a fixed model prompt (Appx. \ref{app:model_prompt}) and produces 10 independent robot artifacts. There is no feedback given to the model and all produced artifacts directly enter the Qualification Stage. The top bot per run is chosen through a round-robin tournament between all the qualified bots.

\textbf{Verifier-Grounded Refinement Harness \textsc{(VGH)}.} In \textsc{VGH} we conceptualize the model's role as that of an optimizer over 10 sequential iterations: the model reads verifier feedback, a journal of prior attempts, and the match feedback, and proposes a better artifact. After each iteration, the harness validates the artifact, qualifies it against the stationary block. The full match feedback (game state at every time step), verifier feedback, design history (a JSON with what it changed on each iteration), and the current best artifact so far is given back to the model. The model can choose to iterate based on the full match feedback on the current best design. Every valid artifact enters a round robin and the Bradley-Terry Elo winner is chosen as the final artifact that enters the Qualification Stage. 

\textbf{Design Lab Harness (DLH).} We conceptualize a model as an open-ended engineer working in a \emph{Design Lab}, a sandboxed workspace containing the simulator and verifier source, the task documentation, and shell and Python execution. The model is given a budget of 10 API calls, which it can use to chain several tool calls that allow it to build, simulate, and analyze match feedback (Table \ref{tab:dlh_tools}). We show an example iteration using \textsc{DLH} from Fable 5.1 in Fig.~\ref{fig:example-refinement-harness}. We use the same process as \textsc{VGH} to choose the final artifact. 

\subsection{Tournament}
\label{sec:methods:tournament}

\textbf{Artifact Leaderboard}. We first run the above \texttt{(model, harness)} for all models defined in Appx.~\ref{app:model_library} and for the three harnesses defined in Sec.~\ref{sec:method_harness}. After each run, a single robot artifact is selected based on harness specific selection method described above (Sec.~\ref{sec:method_harness}). Each artifact is then validated and qualified through matches against a cube. For each (model, harness) pair this is repeated three times yielding three artifacts per model per harness. Artifacts that pass validation and qualification enter the Top-3 robot artifact tournament. These sets of top-3 artifacts play a cross-model round robin for 3 unique seeds (each unique seed plays two matches-one from each starting position so a total of 6 matches). We then compute a Bradley-Terry ELO score to get the top-3 resulting ratings, with a draw counted as half a win, form the \href{https://artifactarena.ai/leaderboard/artifacts}{Artifact Leaderboard}. Using the \href{https://artifactarena.ai/leaderboard/artifacts}{Artifact Leaderboard}, we also compute a \href{https://artifactarena.ai/leaderboard/labs}{Lab Leaderboard} which is used to visualize the range of robot artifacts selected by all the models per research lab. We call this phase of the tournament the Top-3 Round and make all the matches
available \href{https://artifactarena.ai/matches?round=top3}{\texttt{https://artifactarena.ai/matches?round=top3}}.

\textbf{Model Leaderboard}. Our goal is to evaluate models by the best artifact they produce. Therefore, we select the best artifact for every \texttt{(model, harness)} pair from the \href{https://artifactarena.ai/leaderboard/artifacts}{Artifact Leaderboard}. We then play a new cross-model round robin tournament with 11 unique seeds (22 total matches) to compute the top-1 Bradley-Terry Elo score for the model's strongest artifact. This top-1 score is used to create \textsc{ArtifactArena} Champion Leaderboard (Fig.~\ref{fig:main_leaderboard}) also available at \href{https://artifactarena.ai/leaderboard/depthchart}{Champions Leaderboard}. We call this phase of the tournament the \texttt{Champion Round} and make all the matches available \href{https://artifactarena.ai/matches?round=Champion+Round}{\texttt{https://artifactarena.ai/matches?round=Champion+Round}}.

As an independent robustness check, we replayed the 21 frozen model champions generated by their respective harnesses on two disjoint sets of eleven seeds, once from each side. The resulting rankings have a Spearman correlation of $\rho=0.984$ and preserve 201 of 210 (95.7\%) pairwise model orderings, with the top three models unchanged in order (Appx.~\ref{sec:fresh_seed_robustness}).

\textbf{Uncertainty.} Error bars in Fig.~\ref{fig:main_leaderboard} are 95\% bootstrap intervals over generation runs and match outcomes. In each of 1,000 replicates, we resample the three finalists for every model--harness pair and the 22 outcomes for every finalist pair, then refit Bradley--Terry, re-select each harness champion, and refit on champion matches. Since all finalists played each other, this only reweights existing matches. We do not resample within-run finalist selection, which would require bootstrapping every candidate-level round robin, so these intervals are a lower bound on full-pipeline uncertainty.

\begin{figure*}[t]
\centering
\includegraphics[width=\textwidth]{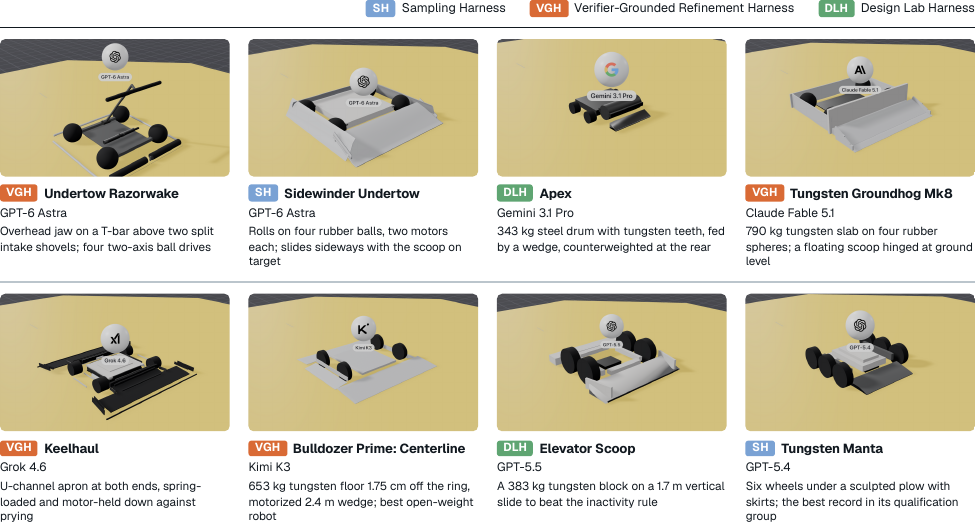}
\caption{\textsc{Visualization of top artifacts.} We render eight robots from the final tournament across the three harnesses each with a different winning strategy. The ELOs of the above artifacts are available here: \href{https://artifactarena.ai/leaderboard/artifacts}{Artifact Leaderboard}.}
\label{fig:visualize_bots}
\end{figure*}

\section{Results}
\label{sec:results}

\subsection{\textsc{ArtifactArena} Leaderboard}
\label{sec:results:leaderbaord}

\noindent\textbf{No one harness fits all:}
Fig.~\ref{fig:main_leaderboard} compares models by the performance of their strongest artifacts across the three harnesses and Fig.~\ref{fig:visualize_bots} shows the top bot morphologies from the models. \textsc{VGH} achieves the highest reported Elo for 15 of the 21 models, including GPT-6 Astra (1539), Claude Fable 5.1 (1310), and Gemini 3.8 Flash (1223), which occupy the first three positions. However, SH produces the strongest artifacts for GPT-5.6 Sol, GPT-5.4, Gemini 3.1 Pro, and GPT-5.5, while \textsc{DLH} does so for Claude Fable 5 and GLM 5.3. We visualize top artifacts in Fig.~\ref{fig:visualize_bots}. 

Despite the possibility of matchup-specific counter-strategies, tournament performance exhibits strong global dominance: in the cross-model round-robin tournament of the top three artifacts from each model–harness pair, the highest-rated artifact which was created by GPT-6 Astra using VGH, wins 91\% of matches.

\textbf{Unreasonable effectiveness of $\mathbf{Sampling Harness (SH)}$.} We find a remarkable ability in both closed- and open-source models to zero-shot fully functioning robot artifacts for our highly open-ended task. Despite receiving no simulation feedback during generation, GPT-6 Astra reaches 1352 Elo with \textsc{SH}, Fig.~\ref{fig:main_leaderboard}, trailing only itself with \textsc{VGH}. Even though iteration leads to more stable bots, we hypothesize that as models develop better world models of 3D or MuJoCo-type environments, harness-engineering for engineering and scientific tasks may be bitter-pilled. We do additional experiments, sampling 250 bots from 21 models to explore zero-shot generation in Sec.~\ref{sec:results_creativity}.

\begin{figure}[!t]
  \centering
  \captionsetup[subfigure]{justification=centering, font=small}
  \begin{subfigure}[t]{0.32\linewidth}
    \centering
    \includegraphics[width=\linewidth]{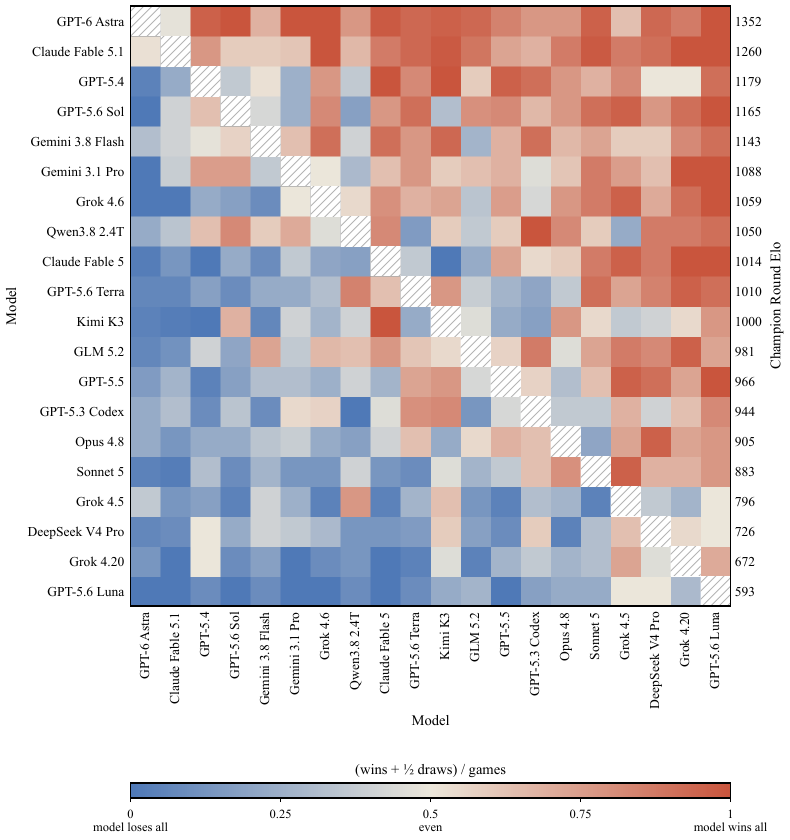}
    \caption{\raggedright\hyphenpenalty=10000 Sampling (SH)}
    \label{fig:winrate-sh}
  \end{subfigure}\hfill
  \begin{subfigure}[t]{0.32\linewidth}
    \centering
    \includegraphics[width=\linewidth]{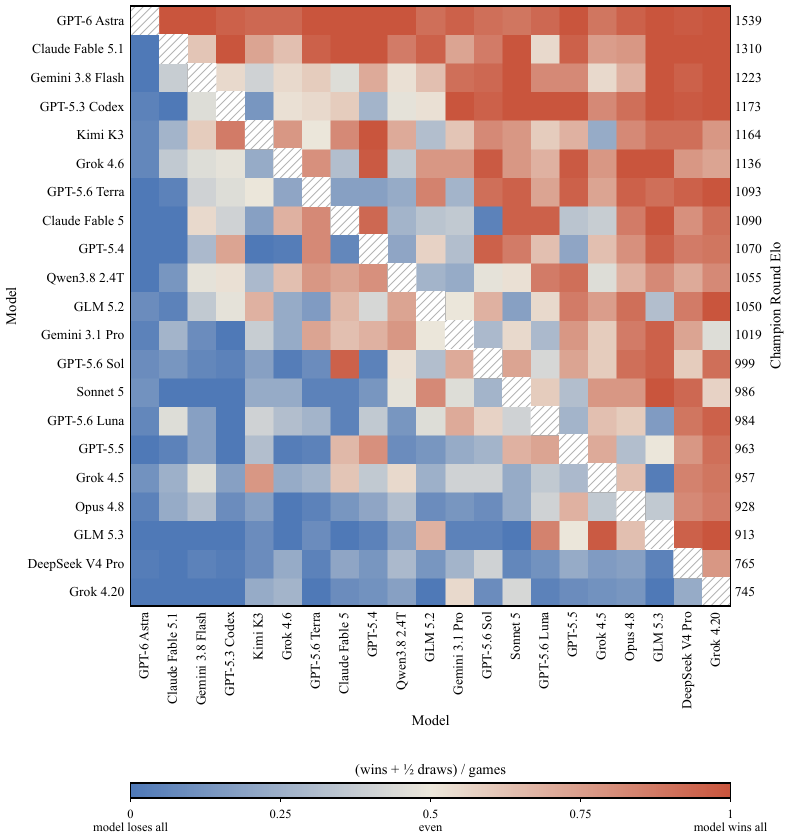}
    \caption{\hyphenpenalty=10000 Verifier-Grounded Refinement (VGH)}
    \label{fig:winrate-vgh}
  \end{subfigure}\hfill
  \begin{subfigure}[t]{0.32\linewidth}
    \centering
    \includegraphics[width=\linewidth]{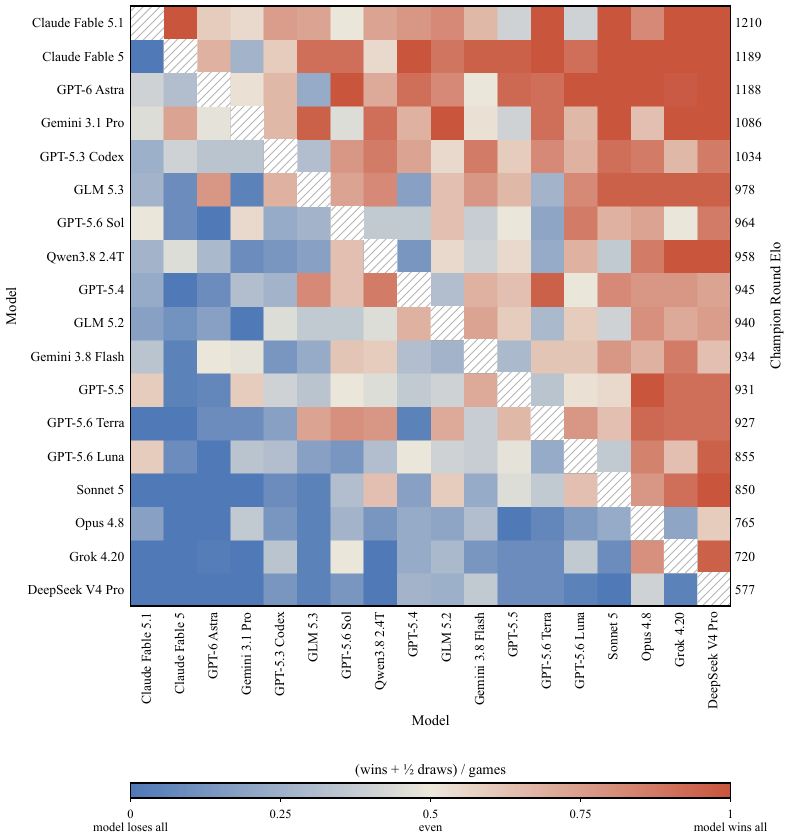}
    \caption{\hyphenpenalty=10000 Design Lab (DLH)}
    \label{fig:winrate-dlh}
  \end{subfigure}
  \begin{flushleft}
  \caption{\textsc{Harness-specific Winrate-Matrices show open-ended harness remain a direction of future work.} Across harnesses, sampling creations are competitive with those created by \textsc{VGH} while none of the top-7 artifacts come from \textsc{DLH} (\href{https://artifactarena.ai/leaderboard/artifacts}{Artifact Leaderboard}). Therefore, we plot the win-rate-matrix to get head-to-head comparisons within each harness and order models by their Elo scores (the top row and left most column are the models with the highest Elo). The uneven color patterns reveal matchup-specific strengths and weaknesses. Within harnesses, we see instability in base model performance as more agency is given: GPT-6 Astra no longer dominates in \textsc{DLH (c)} like it did in \textsc{SH (a)} or \textsc{VGH (b)}, while open source models become more competitive globally with \textsc{VGH}. Complete agency, \textsc{DLH}, also didn't translate to global performance. We believe that developing reliable open-ended design loops that construct stronger competitive artifacts remains a direction for future work. Details in Sec.~\ref{sec:results:leaderbaord} and Appx. Fig.~\ref{fig:winrate-matrices}.}
  \label{fig:pairwise_matrix}
  \end{flushleft}
\end{figure}

\textbf{Growing pains of open-ended design loops.} We find that giving models greater agency over the design loop does not yet consistently lead to stronger artifacts: under the tested same API-call budget, \textsc{(DLH)} trails \textsc{(VGH)} for most models, has no top-7 artifact (\href{https://artifactarena.ai/leaderboard/artifacts}{Artifact Leaderboard}), and produces no champions for three models. We analyze this in Fig.~\ref{fig:pairwise_matrix} by restricting Champion Round matches (described in Sec.~\ref{sec:methods:tournament}) to head-to-head comparisons within each harness and order models by their Elo score (the top-left and bottom left are the ones with the highest score). Each matrix cell reports the row model's score against the column model, $(\mathrm{wins} + \frac{1}{2}\mathrm{ties})/\mathrm{games}$, with red indicating an advantage and blue a disadvantage. The uneven color patterns reveal matchup-specific strengths and weaknesses: higher-ranked artifacts do not uniformly beat lower-ranked opponents. GPT-6 Astra dominates with \textsc{SH} and \textsc{VGH} harness and produces the artifacts that beat the artifacts of every other model produced (darker red squares means model will all matches), however, the same base model ranks third under \textsc{DLH} under Fable 5.1 and Fable 5. Moreover, no single base model dominates the \textsc{DLH}-only category (Fig.~\ref{fig:pairwise_matrix}.c vs. Fig.~\ref{fig:pairwise_matrix}.a). This shows that, for both closed- and open-source models, greater agency over the design loop does not yet consistently yield stronger artifacts when compared with those produced by the same models using simpler workflows under the same API-call budget. We believe that developing reliable open-ended design loops that construct stronger competitive artifacts remains a direction for future work.

\subsection{Emergent Creativity in ArtifactArena}
\label{sec:results_creativity}

\textsc{ArtifactArena} reveals creativity in both robot design and competitive behavior: models develop attack and defense mechanisms, coordinate hardware with control, and use physical interactions and game rules to gain an advantage. Under the same API-call budget, we now examine how these capabilities appear across the sampling, refinement, and open-ended harnesses. 

\noindent \textbf{Hardware creativity.} Top models co-design hardware-software to implement the chosen attack, defense, and recovery strategy. From the 509 buildable robots in the sampled subset, 208 robots had powered scoops or plows, 31 with flippers, 21 with spinners or drums, ten with clamps, and three with movable weights. GPT-5.5 describes \emph{Keelback Twin-Scoop} as having ``active front/rear scoop plates so it can attack in either direction.'' GLM-5.3 gives \emph{Duozer Mk VII} a foam dome, explaining that ``a climber gets a single low-friction contact point on a convex surface and slides off.'' We find models implement creative ways to exploit the rules including movable weights to increase COM height to avoid getting pinned. See Appx.~\ref{app:hardware_analysis} for more details.

\noindent  \textbf{Controller creativity.} Models write controllers as complex rule-based state machines that respond to the opponent's design, movement, and the arena rules: in the refinement runs, 63\% of controllers organize their behavior into distinct modes with explicit conditions to switch between them. Across the controller corpus, we identify 209 distinct mode names, including \texttt{chase}, \texttt{flank}, \texttt{dump}, \texttt{wiggle}, \texttt{wheelie}, \texttt{holdout}, and \texttt{activity\_win}. Controller examples range from 71 lines for Grok 4.20 to 1,700 lines for GPT-5.4. GPT-6 Astra checks the opponent's motors for potential weapons before choosing to intercept the opponent instead of directly engaging. GLM-5.3's controller classifies opponent morphologies as spinners, flyers, or robots that stay near the edge and implements custom behavior against each. Kimi K3's controller holds a pin when the opponent's inactivity timer will expire first and implements a sharp breaking behavior near the edge so that the opponent's momentum can carry it off the platform. Claude Fable 5.1 uses contact measurements to estimate the weight supported by each wheel and limits motor torque according available rubber wheel grip. See Appx.~\ref{app:controller_analysis} for more details.

\begin{figure}[!t]
    \centering
    \includegraphics[width=0.95\textwidth]{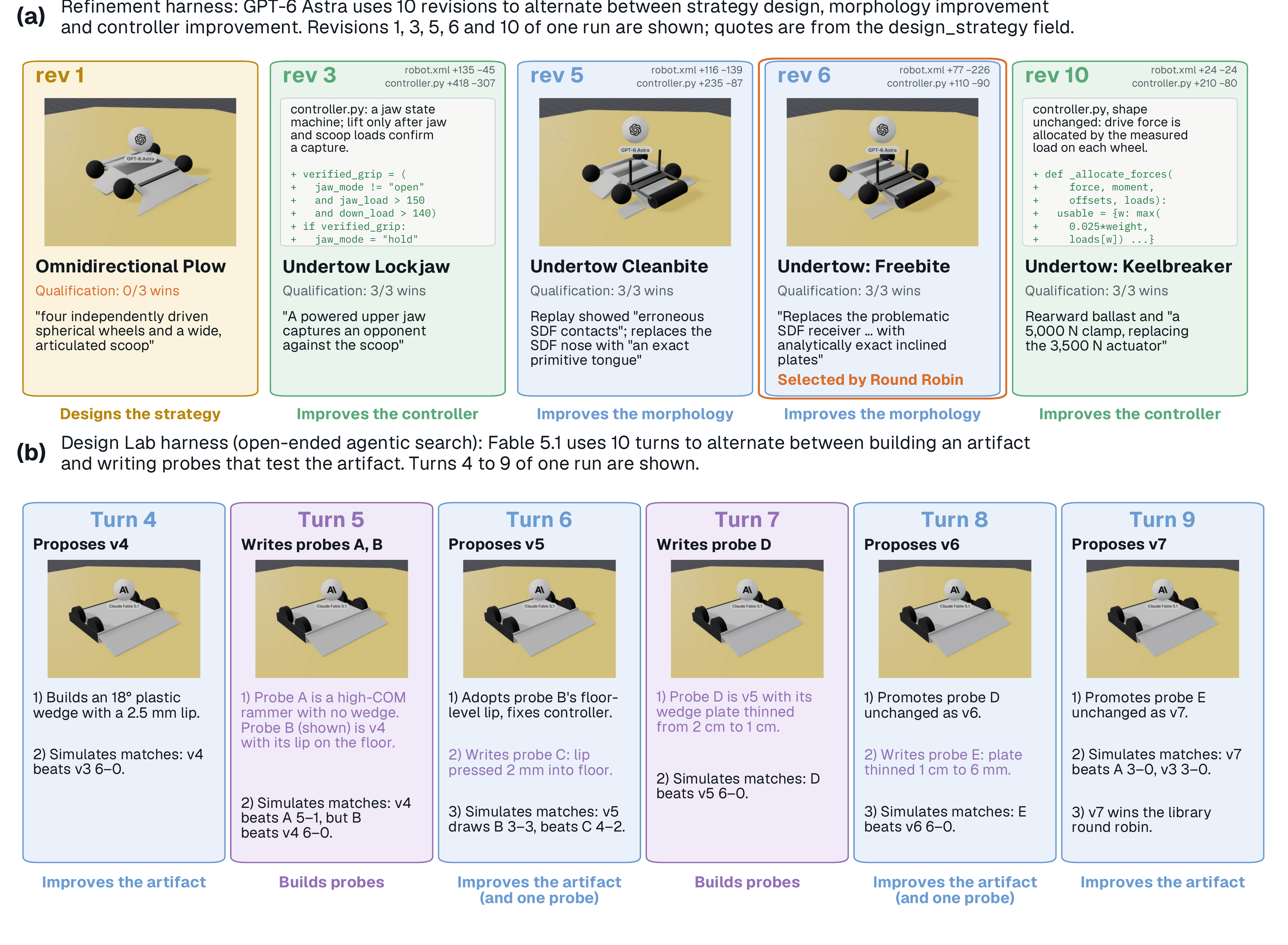}
    \caption{\textbf{An example refinement vs. design lab run.} (a) Example refinement by GPT-6 Astra: box's are colored by the largest change (by lines added and removed in robot.xml and controller.py): strategy (gold), morphology (blue) or controller (green). Revision 6, outlined in orange, is the robot the run's round robin selected for the tournament. (b) Claude Fable 5.1's turns in the Design Lab iterate between building an artifact (blue) and making probes to test the artifact (purple).}
    \label{fig:example-refinement-harness}
  \vspace{-5mm}
    
\end{figure}

\textbf{Zero-shot design creativity.} Given the remarkable emergent zero-shot design capabilities of frontier models in Mujoco, we generated 250 samples per model for 21 models. Across 5,250 total API calls, 4,384 artifacts are built and 2,242 qualify: Claude Fable 5 qualifies 91\% of its samples, Claude Fable 5.1 84\%, Gemini 3.8 Flash 76\%, GPT-5.5 71\% and GPT-5.6 Sol 69\%; GPT-5.6 Luna qualifies 24\%, Claude Sonnet 5 33\%, and Gemini 3.1 Pro, whose champion won the cross-model tournament, only 40\%; DeepSeek V4 Pro qualifies 4\%, Grok 4.5 and Grok 4.20 2\%. Even though zero-shot design is a significantly harder task, models surprisingly use the full design space: 29\% of robots include a custom mesh, 192 drive motors through tendons, 253 use passive springs, 171 use ball joints and 91 use slide joints, and the largest reach 15 bodies, 58 geometries and 14 motors. 

\subsection{Remarkable capabilities that surprised us}

Several behaviors appeared that surprised us. For example, models designed controllers that plan for accidentally being flipped zero-shot, anticipating this issue without any feedback or experience from the arena. In the \textsc{sampling harness}: 96 controllers detect inversion from the root orientation and remap their drive, with wheels built to stand taller than the hull. Claude Fable 5.1's \emph{Groundhog Tungsten} carries a lift angle ``that puts the plate on the ground when we drive upside down'': \Verb|inverted = upz < -0.2; sgn = -1.0 if inverted|. In \textsc{VGH}, 17 controllers do the same, and Gemini 3.1 Pro's \emph{Apex Juggernaut} reverses its drum when inverted so the exposed surface still strikes upward: \Verb|is_upside_down = abs(my_roll) > 2.0 or abs(my_pitch) > 2.0; drive_fwd = -drive_fwd; weapon_cmd = 1.0|.

Some controller designs create strategies tracking the  difference in their and their opponents inactivity timers, raising their center of mass in the final seconds of a mutual stall to win the height tie-break. Moreover, some controllers also estimate physics from their observations by reconstructing the normal load on each wheel from contact impulses, while some predict interception points. GPT-6 Astra (\textsc{VGH}) inspects its opponents morphology and treats any opponent hinge above 1,800 N$\cdot$m as a weapon before choosing its approach. 

Lastly, in \textsc{DLH}, models alternate between building the artifact and designing strategies that can beat it. Fig.~\ref{fig:example-refinement-harness}.B shows Claude Fable 5.1 saved its first design ``as a sparring baseline,'' then wrote two probe opponents to attack its own wedge, a ``vertical-bumper high-COM rammer on the same chassis'' and a copy ``with wedge lowered 2 mm so its lip touches the platform''.

\subsection{What makes some models creative?}
\label{sec:results_what_makes_models_creative}

\textsc{ArtifactArena}'s simple rules and vast, physically grounded design space give models wide freedoms to engineer robots, therefore we ask what makes some models much better at producing novel high-Elo artifacts? We find that the most creative models are creative across two complementary dimensions: 

\textbf{Creative strategic decomposition of task.} Creative strategic decomposition asks for models to turn the broad goal of ``best possible bot'' into smaller objectives that guide their designs. These often require reasoning at a higher level, decomposing the strategy into sub problems that can be independently tested and implemented as verifiers in the simulation. Claude Fable 5.1's \emph{Groundhog Tungsten} begins with a strategy of ``getting its blade under the opponent'' and ``lifting to transfer their front-axle weight onto our front tyres,'' followed by pushing the opponent toward the edge and ``dumping them by lowering the arm and reversing.'' Fable 5.1 further decomposes this into problems that the controller must solve to implement each strategy into state machines that adjust the lift angle according to opponent mass and state machines that limit speed near the edge. Top models are able to decompose non-quantifiable goals into a series of strategies and then concrete and verifiable sub-goals that the hardware-software co-design must meet. 

\textbf{Creative physical implementation of strategy.} Models must also understand how to use the physically grounded design space to achieve their chosen strategies. Creativity here lies in translating a strategy into effective robot morphology and control, as a useful strategy is only as good as how its implementation works under the simulated physics. Across refinement runs, models often retained their strategies while changing their physical implementation. For example, Kimi K3's strategy was ``a low, near-maximum-mass wedge that gets under the opponent and shoves it off the platform''. 
Its first five implementations were four-wheel skid-steer slabs which only won two out of fifteen qualification rounds. At iteration 6, it finally identified how to change the morphology and controller to implement that design by replacing it with a four-wheel wedge with a center-pivot design. This improved qualification performance to 3-0 in the qualification stage. Such creative implementations is seen across models and are detailed in the Appendix~\ref{app:physical-implementation}.

\textbf{Role of refinement in model creativity.} The harnesses reveal different aspects of model creativity within these dimensions. \textsc{SH} produces a median of 19 structurally distinct designs, compared with 12 for verifier-grounded refinement (across iterations). During refinement, models mostly stuck to their initial strategies and used the feedback for fixes. Claude Fable 5, for example, identified an edge overrun in its trajectory and replaced a fixed edge threshold with a braking-distance speed limit. Across the refinement runs, 19 of 21 models reach a qualifying design in every run within ten revisions. We anticipate harnesses that allow models to work across these creative dimensions to be the most successful. We also briefly discuss ambitious designs that ultimately failed in Appx.~\ref{sec:ambitious_designs}.

\subsection{Limitations} 
All simulations are wrong, but some are useful. The main limitation of \textsc{ArtifactArena} is the gap between simulation and reality. Robots are built and judged inside MuJoCo therefore constructing a robot here doesn't mean that the robot is going to work in the real world. Nevertheless, we anticipate that as simulations get closer to reality, the same arena and the same harnesses will carry over unchanged: the robot becomes something that can be printed, assembled and fought for real. Therefore, our results should be interpreted as measuring competitive physical design and control of frontier models in simulation rather than direct real-world robotic capability. 

\section{Related Work}
\label{sec:related}

\textbf{Prescribed objectives and executable verification.} MMLU~\citep{hendrycks2021mmlu} and MMLU-Pro~\citep{wang2024mmlu} test question answering, while HumanEval~\citep{chen2021humaneval}, SWE-bench~\citep{jimenez2024swebench}, and LiveCodeBench~\citep{jain2024livecodebench} test program correctness, and KernelBench~\citep{ouyang2025kernelbenchllmswriteefficient} measures execution speed. WebArena~\citep{zhou2024webarena} and BEHAVIOR-1K~\citep{pmlr-v205-li23a} extend evaluation to interactive web environments and simulated everyday activities, respectively. These benchmarks use prescribed task objectives. Chatbot Arena~\citep{chiang2024chatbotarena} instead uses pairwise human preferences, while judgment-based evaluation can exhibit position and verbosity biases~\citep{3666122.3668142}. SWE-Arena~\citep{zhao2025swearenainteractiveplatformevaluating} evaluates multi-round software engineering, while CodeClash \citep{yang2025codeclashbenchmarkinggoalorientedsoftware} evaluates competing codebases refined through match feedback. \textsc{ArtifactArena} extends executable competition to physically simulated robots, with models designing both their morphology and controller.

\textbf{Physical reasoning and functional construction.} Physically grounded benchmarks test capabilities ranging from action selection in PHYRE \citep{NEURIPS2019_4191ef5f}, to outcome prediction in Physion \citep{bear2022physionevaluatingphysicalprediction}, physical understanding in PhysBench \citep{chow2025physbench}, and experimental discovery in PhysGym \citep{chen2025physgym}. Construction benchmarks go further by making model output a physical system. BuildArena is a close example, evaluating language-driven construction through simulated transport, support, and lift tasks \citep{xia2026buildarena}. BesiegeField similarly evaluates machine programs assembled from standardized components, refined through feedback \citep{zhang2026compositionalmachinedesignprogram}. \textsc{ArtifactArena} shares their emphasis on constructing functioning systems, while evaluating generated robots through direct physical competition in a non-saturating setting. 

\textbf{Co-design and evolving competition.} Prior work jointly optimizes bodies and behavior through competitive evolution~\citep{sims, gupta2021embodied, doi:10.1126/sciadv.ady2888}. Evolution Gym benchmarks body and controller optimization~\citep{bhatia2021evolution}, and RoboMoRe uses LLMs to co-design morphology and reward~\citep{fang2025robomorellmbasedrobotcodesign}. Competitive reinforcement learning produces complex behavior in simulated sumo environments~\citep{bansal2018emergentcomplexitymultiagentcompetition}, while Enhanced POET evolves environments and solutions together~\citep{pmlr-v119-wang20l}. In \textsc{ArtifactArena}, new opponents expose weaknesses in existing designs and raise the bar for future submissions.


\section{Acknowledgments}
This work was supported by the DARPA Mathematics for the DIscovery of ALgorithms and Architectures (DIAL) program, the DARPA Mathematics of Boosting Agentic Communication (MATHBAC) program, the NSF award 2124052, the MIT Generative AI Impact Consortium, the MIT-IBM Watson AI Lab, the Department of the Air Force Artificial Intelligence Accelerator under Cooperative Agreement Number FA8750-19-2-1000, and the Air Force Office of Scientific Research (AFOSR) under award number FA9550-21-1-0399.

The views and conclusions contained in this document are those of the authors and should not be interpreted as representing the official policies, either expressed or implied, of the Department of the Air Force or the U.S. Government. The U.S. Government is authorized to reproduce and distribute reprints for Government purposes notwithstanding any copyright notation herein.

\clearpage
\bibliographystyle{plainnat}
\bibliography{iclr2027_conference}

\clearpage
{\LARGE \textbf{Appendix}}
\FloatBarrier
\section{AI Statement}

In this work, we used generative AI tools to assist with research execution, including implementing methods and writing code, running experiments, and analyzing and interpreting experimental results. We additionally used these tools to create scientific figures and correct grammar. The models evaluated in ArtifactArena generated a synthetic dataset of robot artifacts, comprising robot morphologies and controllers, as part of the experimental procedure described in the paper. We did not use generative AI to generate the core research ideas or draft the substantive scientific content of the manuscript. However, we used generative AI tools to create scientific figures, correct grammar. We reviewed all AI-assisted contributions, checked code through execution and testing, verified analyses and figures against the underlying experimental outputs. We take full responsibility for the final content of this work, including its text, methods, results, claims, and artifacts produced with the assistance of generative AI.

\section{Results (continued)}

\subsection{Fresh-seed robustness}
\label{sec:fresh_seed_robustness}

To test whether the leaderboard is sensitive to the particular tournament seeds, we froze the 21 champion artifacts selected by the original tournament and replayed their round robin on two disjoint sets of eleven seeds: the reference seeds and a held-out set. Each seed was played once from each side, yielding 4,620 matches per set (9,240 total). Both sets were rerun using the same current harness to isolate seed sensitivity from changes to the runtime, and no artifacts were selected using held-out results. The resulting model rankings were highly consistent (Spearman $\rho=0.984$, 95\% bootstrap interval $[0.948, 0.990]$); 201 of 210 pairwise model orderings (95.7\%) were preserved, seven reversed, and two became ties. The top three models were unchanged in order. For the Elo and rank-correlation intervals, we use 2,000 seed-block bootstrap replicates, independently resampling each pair's eleven seed blocks within each set while keeping the two side-swapped matches from each seed together.

\subsection{Ambitious Designs} 
\label{sec:ambitious_designs}

Failure is critical to creativity and we briefly discuss some of the most ambitious designs in the set that never won a match, but mark the frontier of what a model will attempt: Gemini 3.1 Pro's \emph{Centrifuge}, a 188 kg vertical drum spinner whose reaction torque backflipped its own chassis; GPT-5.6 Terra's \emph{Orbitbreaker}, a 0.68 m aluminum flywheel with eight impact teeth; Claude Opus 4.8's \emph{Tungsten Tortoise}, which slings a steel slab below its axles as a self-righting pendulum; GPT-5.5's \emph{Keeljaw Sumo Loader}, whose intake drum drags opponents up its ramp; and nine of the sixteen sphere-drive lifters GPT-6 Astra placed in the set.

\subsection{Artifact Morphology Analysis}
\label{app:hardware_analysis}

We analyze a subset of 630 sampled artifacts from 21 models. We report the mechanisms present in these artifacts and use excerpts from the models' design rationales to describe their intended functions. Qualification and match results provide separate evidence of how the completed robots perform.

\textbf{Powered mechanisms.} 208 robots had powered scoops or plows, 31 with flippers, 21 with spinners or drums, ten with clamps, and three with movable weights while ten robots carry weapons at both ends. These mechanisms provide different ways to engage an opponent, including lifting, striking, gripping, and shifting weight.

\textbf{Attack.} Several models design mechanisms intended to lift opponents and reduce their grip on the platform. GPT-5.6 Sol's \emph{Gravemouth Leviathan} combines a wedge with a rubber intake drum. Its rationale states that the drum ``rotates upward at its exposed front face and rearward over its top, lifting an opponent while drawing it onto the wedge.'' Gemini 3.1 Pro's \emph{OblivionScoop} combines a scoop with a powered jaw that, according to the model, ``bites down on the opponent.'' Other designs allow attacks from either end: Grok 4.6's \emph{Janus} has front and rear scoops, while GPT-5.5 describes \emph{Keelback Twin-Scoop} as using ``active front/rear scoop plates so it can attack in either direction.''

\textbf{Weight distribution.} Some models use movable weights to change how their robots push and recover. GPT-5.4's \emph{Shiftplow} moves its weight according to the situation; the model states that it ``throws that ballast forward during charges and pushes, back during edge escape and turning.'' Its refinement design \emph{Centerline Kraken} moves a 553\,kg core along two axes. The rationale describes this choice as ``controllable leverage.'' GPT-5.3-Codex's \emph{Wedgewarden Atlas} instead raises a tungsten weight ``when both robots are close to inactivity timeout, increasing COM height.'' This design targets the rule that favors the higher center of mass when both robots lose to inactivity on the same step.

\textbf{Defense and recovery.} Models also design hardware intended to prevent opponents from getting underneath them or to preserve movement after being overturned. GPT-5.4's \emph{Tungsten Manta} includes ``side skirts and a rear skid to reduce easy side entries and make reverse escapes survivable.'' GPT-5.6 Sol's \emph{Tungsten Undertow} leaves its wheels exposed so that, according to its rationale, they ``remain able to propel the robot even if it is inverted.'' GLM-5.3's \emph{Duozer Mk VII} uses a foam dome intended to shed climbing opponents: ``a climber gets a single low-friction contact point on a convex surface and slides off.'' Its \emph{Widowmaker} uses one flipper for both attack and recovery, with the stated purpose of being able to ``throw opponents and self-right the robot.''

\textbf{Hardware changes under refinement.} Defensive features appear more frequently in refinement than in the sampled comparison set. The reported share of robots with skirts increases from 29\% to 51\%, and domed roofs from 3\% to 7\%. Individual histories show models adding and changing mechanisms. Claude Fable 5.1's first run begins with the single-ended \emph{Groundhog} and adds a second ramp at revision four as \emph{Janus Anvil}. The revised robot also carries a swinging 320\,kg tungsten weight, which the model describes as shifting its center of mass ``to win stalls, timers and tiebreaks.'' Every revision from the fourth onward qualifies 3--0, and revision nine is selected with a 38--16 record in the run's selection round robin.

\textbf{Hardware failures.} Intended advantages do not always translate into functioning designs. Nine of GPT-6 Astra's sixteen sampled sphere-drive lifters fail qualification, mostly through size violations or inactivity. Its \emph{Omnidirectional Traction Thief} has spheres that, according to the rationale, ``protrude above and below the fixed chassis'' for ``bidirectional survivability.'' However, it loses all three qualification rounds because its dimensions exceed the allowed box during movement.


\subsection{Controller Analysis}
\label{app:controller_analysis}

Controllers determine how robots use their hardware during a match. They receive observations of both robots and the arena, allowing them to respond to the opponent's movement, contact with other surfaces, and the conditions that determine a loss. We examine how models use this information to choose tactics and control their robots.

\textbf{Decision structure.} In the refinement runs, 63\% of controllers organize their behavior into distinct modes and use explicit conditions to switch between them. Across the controller corpus, we identify 209 distinct mode names, including \texttt{chase}, \texttt{flank}, \texttt{dump}, \texttt{wiggle}, \texttt{wheelie}, \texttt{holdout}, and \texttt{activity\_win}. Controller examples range from 71 lines for Grok 4.20 to 1,700 lines for GPT-5.4.

\textbf{Responding to the opponent.} Models use the opponent's observable hardware to choose how to engage. GPT-6 Astra inspects the opponent's motors and treats a hinge with torque above 1,800\,N$\cdot$m as a potential weapon before choosing an approach. It also calculates where to intercept the opponent rather than directing the robot toward the opponent's current position. GLM-5.3 classifies opponents as spinners, flyers, or robots that stay near the edge. Against a flying opponent it cannot reach, the controller patrols the center and relies on the opponent's inactivity timer to determine the outcome.

\textbf{Using the game rules.} Several controllers choose actions based on which robot is closer to losing through inactivity. Claude Fable 5.1 holds still when the opponent stopped first, retreats when the opponent moves, and raises its mast while holding position to target the center-of-mass height tie-break. Kimi K3 maintains a pin when the opponent's inactivity timer will expire first. These decisions make the timing of movement part of the competitive strategy: releasing an opponent or continuing to push is not always preferable to holding position.

\textbf{Controlling grip and avoiding the edge.} Controllers also account for the physical limits of their hardware. Claude Fable 5.1 uses contact measurements to estimate the weight supported by each wheel and limits motor torque according to the grip available from the rubber wheels. Grok 4.6 calculates edge clearance using the robot's rectangular outline, accounting for corners that could extend beyond the platform. Kimi K3 brakes near the edge with the aim of letting the opponent's momentum carry it off while keeping its own robot on the platform.

\textbf{Responding to inversion.} Some controllers change their commands when the robot is overturned. Gemini 3.1 Pro reverses its drum when inverted so that the exposed surface continues to strike upward. GPT-5.6 Sol's \emph{Tungsten Undertow} similarly detects inversion and reverses its wheel commands. These behaviors complement hardware designed to remain useful in more than one orientation.

\subsection{Elo Calculation}
\label{app:elo}

We report Bradley--Terry (BT) strengths on the Elo scale. Artifact $i$ has rating $R_i$, and the
probability that $i$ beats $j$ is
\begin{equation}
  P(i \succ j) \;=\; \frac{1}{1 + 10^{-(R_i - R_j)/400}}
  \;=\; \sigma\!\left(\theta_i - \theta_j\right),
  \qquad
  R_i \;=\; 1000 + \frac{400}{\ln 10}\,\theta_i ,
\end{equation}
where $\sigma$ is the logistic function and $\theta_i$ is the BT log-strength. Thus a 400-point gap
corresponds to 10:1 odds.

Each game counts as a win (1), draw ($\tfrac{1}{2}$) or loss (0), with both
starting sides included. Because BT strengths are identifiable only up to an additive constant, we
report the maximum a posteriori estimate under an independent prior $R_i \sim \mathcal{N}(1000, 800^2)$.
At this optimum the mean rating of every fit is exactly 1000, which fixes the offset.

\clearpage
\subsection{Tournament results}

\begin{figure}[H]
  \centering
  \captionsetup[subfigure]{justification=centering, font=small}

  \begin{subfigure}[t]{0.60\linewidth}
    \centering
    \includegraphics[width=\linewidth]{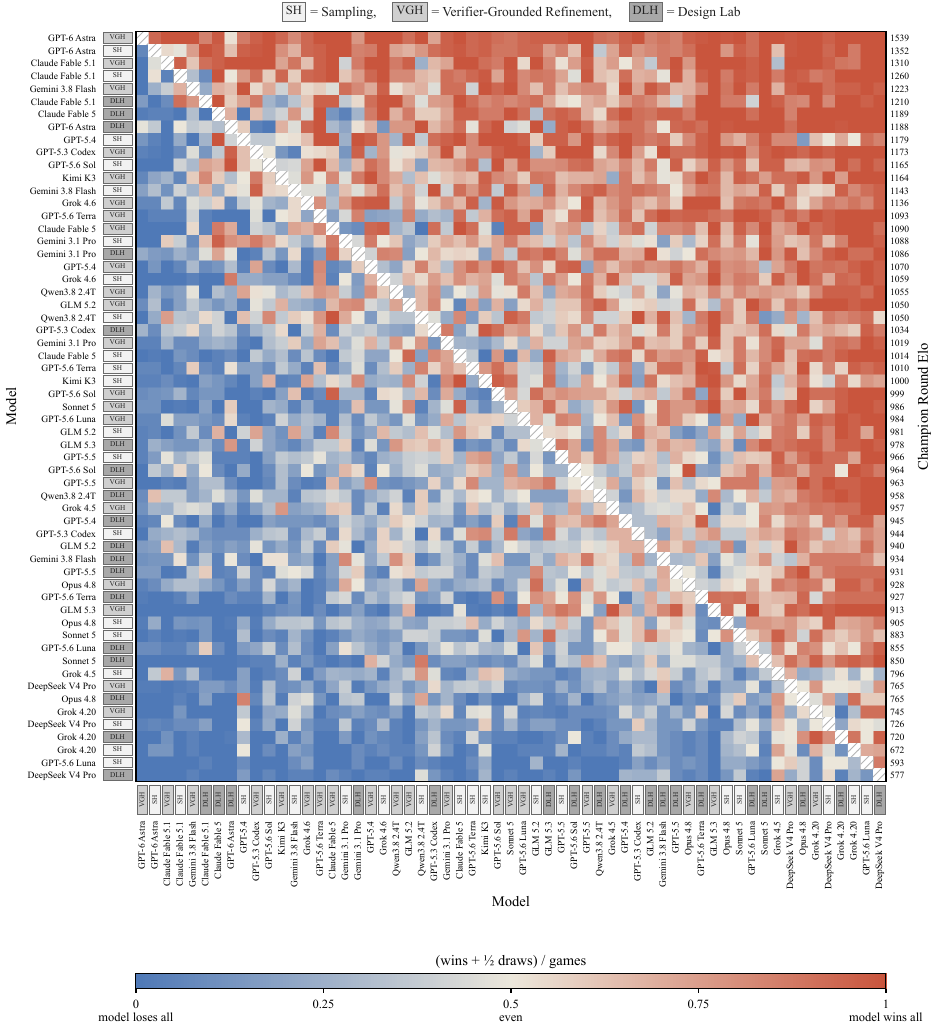}
    \caption{All harnesses}
    \label{fig:winrate-all}
  \end{subfigure}

  \vspace{1em}

  \begin{subfigure}[t]{0.30\linewidth}
    \centering
    \includegraphics[width=\linewidth]{figures/appx_champion_winrate_matrix_sh.pdf}
    \caption{\raggedright\hyphenpenalty=10000 Sampling (SH)}
    \label{fig:winrate-sh}
  \end{subfigure}\hfill
  \begin{subfigure}[t]{0.30\linewidth}
    \centering
    \includegraphics[width=\linewidth]{figures/appx_champion_winrate_matrix_vgh.pdf}
    \caption{\hyphenpenalty=10000 Verifier-Grounded Refinement (VGH)}
    \label{fig:winrate-vgh}
  \end{subfigure}\hfill
  \begin{subfigure}[t]{0.30\linewidth}
    \centering
    \includegraphics[width=\linewidth]{figures/appx_champion_winrate_matrix_dlh.pdf}
    \caption{\hyphenpenalty=10000 Design Lab (DLH)}
    \label{fig:winrate-dlh}
  \end{subfigure}

  \caption{\textbf{Champion Round head-to-head results.}
  Each matrix shows every pairing of cell champions (the best artifact of each model under each harness).
  Cell $(i,j)$ is the score of the row model against the column model,
  $(\text{wins} + \tfrac{1}{2}\,\text{draws}) / \text{games}$
  cell $(j,i) = 1 - $ cell $(i,j)$, and the hatched diagonal marks self-pairings, which are never played.
  Models are ordered by Champion Round Elo (right axis)
  \textbf{(a)}~All 59 champions; grey boxes give each champion's harness.
  \textbf{(b--d)}~The same results restricted to pairings within a single harness.}
  \label{fig:winrate-matrices}
\end{figure}

\clearpage

\subsection{Dichotomous Key}

\begin{center}
  \includegraphics[
    width=\textwidth
  ]{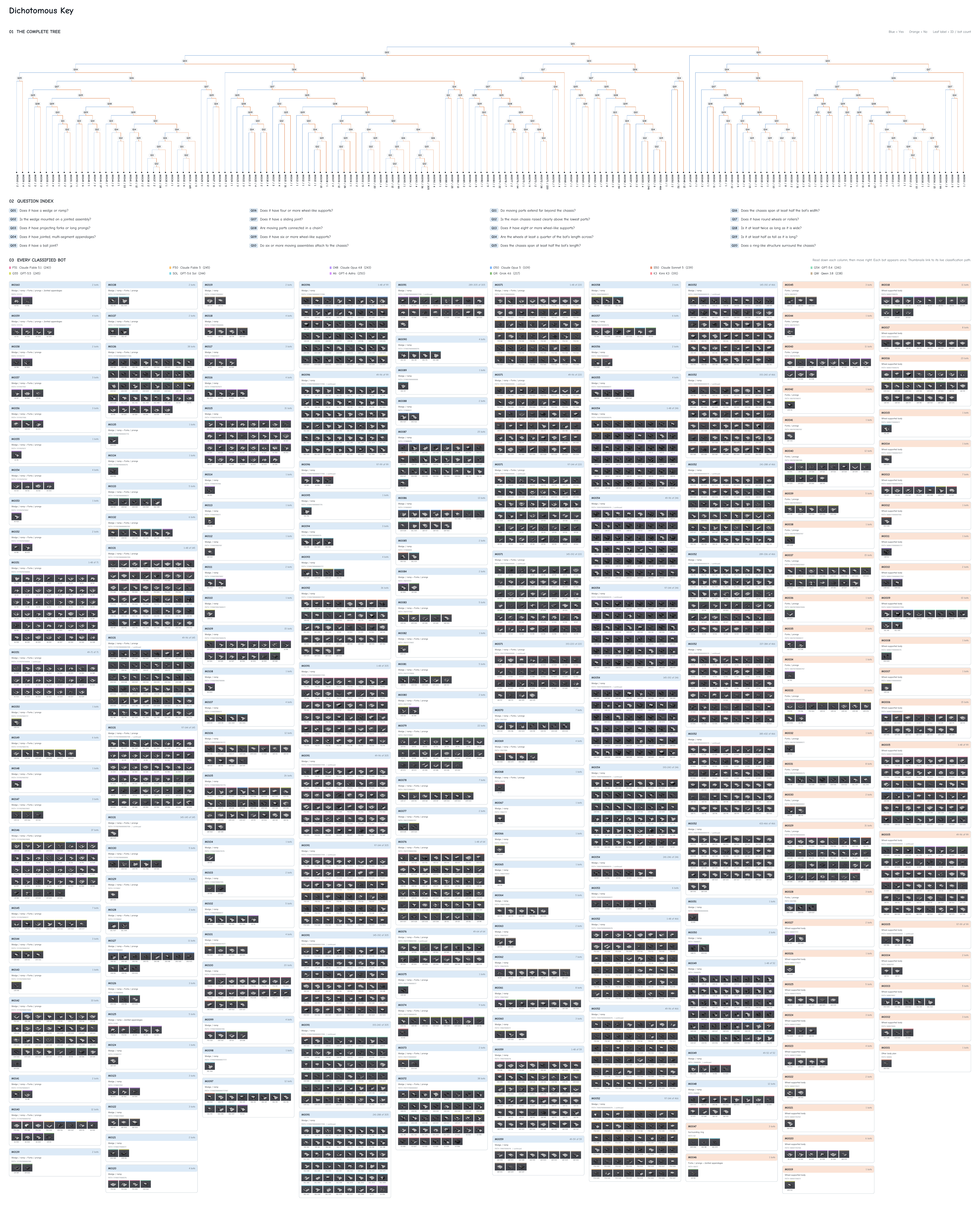}

  \captionof{figure}{Dichotomous key of SH generated bots}
  \label{fig:dichotomous-key}
\end{center}

\clearpage

\subsection{Physical implementation across revisions}
\label{app:physical-implementation}

Some examples of iterating on hardware implementation: Kimi K3 attributed its wedge's poor turning to wheel scrub torque exceeding drive torque. Revision 6, \emph{Fulcrum}, placed the drive weight on two wheels at the center of mass, allowing it to pivot in place while preserving the wedge-pushing strategy. Claude Fable 5.1 initially lost two of three rounds to size violations when its lifter rose. It inferred that the size check used orientation-dependent bounding boxes and repositioned the wheels and ramp, qualifying 3-0 from revision 2 onward. GPT-5.4 replaced a successful pusher's fixed tungsten slab with sliding ballast so the controller could shift mass toward the active pushing end or toward the platform center near an edge.

Implementation failures were common among sampled robots: of 216 robots that built but did not qualify, 125 drove themselves off the platform, 60 triggered the inactivity rule, and 29 exceeded size limits during motion.

\section{Methods (Continued)}
\label{app:harnesses}

\subsection{Detailed Robot Design Constraints}
\label{app:robot_design_details}

\textbf{Structure.} A robot must have exactly one free joint, on its root body, so no part of a robot may detach during a match. Every geometry and every motor must be named, since the controller addresses actuators by name, and each jointed body must carry at least one geometry. Every body, geometry, joint, actuator, and tendon must also carry a short natural-language description of its purpose, which the arena strips before compilation and retains as part of the design rationale. Plain mesh geometry, \texttt{<option>} and \texttt{<contact>} blocks, and MJCF composition elements such as \texttt{<include>}, \texttt{<replicate>}, \texttt{<attach>}, \texttt{<composite>}, and \texttt{<flexcomp>} are not permitted. Custom meshes must be closed, outward-facing, and defined directly in the submitted XML, so robots cannot rely on external geometry files; the mass of a signed distance field geometry is computed from the volume enclosed by its mesh.

\textbf{Materials.} The material palette consists of foam, plastic, rubber, carbon fiber, aluminum, titanium, steel, and tungsten. Explicit mass, density, and inertial specifications, as well as gravity compensation, are removed before validation. Total robot mass is geometry mass plus actuator mass.

\textbf{Actuation.} Only motor actuators are permitted; position, velocity, and other MuJoCo actuator types are rejected. A submission cannot directly actuate its root free joint, anchor an actuator to the world, or connect a mechanism to the opposing robot. Each unit of gear magnitude adds 0.01 kg of actuator mass to the body that mounts the motor; for motors on fixed tendons the priced gear is scaled by the sum of the tendon's joint coefficients.

\textbf{Passive Elements.} Damping and friction loss on internal joints and tendons are capped at 600 per degree of freedom and carry no mass cost. Spring stiffness is unbounded, but springs must be unloaded in the initial pose, and armature is fixed at zero. The root free joint cannot carry damping, friction loss, or stiffness, and tendons and actuator transmissions must remain entirely internal to the robot.

\textbf{Contact Model.} Collision filtering, solver parameters, friction settings, and related contact properties are controlled by the arena. Robot geometries receive a higher contact priority than the platform, so the material's own friction coefficient governs traction rather than the platform's; torsional and rolling friction are fixed at 0.005 and 0.0001 for all contacts.

\textbf{Size Enforcement.} The bounding box is measured on the initial pose along the world axes and re-checked in the robot's own root frame at every control step of a match; a robot whose extents exceed the box at any step loses the match, so articulated mechanisms must be designed to stay within these dimensions while moving.

\textbf{Controller Sandbox.} The controller runs in a restricted Python sandbox with only \texttt{math}, \texttt{numpy}, \texttt{random}, \texttt{typing}, and \texttt{collections} available; file, process, and network access are forbidden. Controller state persists across calls within a match and is reset before every match. A controller exception or a malformed return counts as a loss.

\textbf{Validation.} Submissions are validated before qualification: the XML must compile in MuJoCo, every constraint above is checked, and the controller is invoked once on a real observation. A robot that fails any check forfeits the tournament.

\subsection{Design Lab Harness Details}

\paragraph{Loop.}
A model turn is a single API call. Each turn the prompt is rebuilt from durable state rather than accumulated: the system prompt (rules, robot requirements, controller interface, observation schema, and the tool reference below), the static documentation, the current draft (\texttt{robot.xml} and \texttt{controller.py} at the workspace root), a one-line-per-step journal of previous tool calls and their outcomes, the model's own \texttt{notes.md}, and the results of the previous turn's tool calls. The model does not receive the full prior conversation but can write to \texttt{notes.md}. A turn ends with one or more fenced JSON tool calls, each carrying a \texttt{tool}, \texttt{args} and a short \texttt{note}. All calls in a turn run in order and every result is returned. Tool calls do not count against the budget and model can chain tool calls together. The default budget is 10 turns, or fewer if the model calls \texttt{finish}.

\paragraph{Bot references.} Tools that act on a bot accept \texttt{draft} (the working files), \texttt{stationary} (the qualification block, the only built-in opponent), or
the name of a bot the model saved with \texttt{save\_bot}. A match may pit any two references against each other, including the same bot on both sides.

\paragraph{Tools.} The model has eighteen tools, listed in Table~\ref{tab:dlh_tools} in the four groups we use in the main text: workspace, arena, analysis, and library.

\begin{table}[h]
\centering
\small
\begin{tabular}{@{}llp{8.6cm}@{}}
\toprule
Group & Tool & What it does \\
\midrule
Workspace & \texttt{list\_dir} & List a workspace directory, including the bundled simulator source. \\
 & \texttt{read\_file} & Read a file, with line offset and limit for paging through long files. \\
 & \texttt{grep} & Search files under a path for a pattern. \\
 & \texttt{write\_file} & Write a file (robot, controller, notes or experiment script), replacing its contents. \\
 & \texttt{run\_python} & Run inline code or a script in the workspace; the simulation package \texttt{mjarena} is importable. \\
 & \texttt{run\_bash} & Run a shell command in the workspace. \\
\midrule
Arena & \texttt{does\_bot\_verify} & Static validation of morphology and controller, no simulation. Returns pass/fail per part and feedback. \\
 & \texttt{does\_bot\_qualify} & Validate, then run the qualification trials against the block. Returns validation and qualification results with feedback. \\
 & \texttt{run\_match} & Verify both bots, then play $n$ seeds between two references. Qualification is not required. Returns per-seed outcomes, a downsampled replay, and a \texttt{match\_id}; the full recording stays on disk. \\
 & \texttt{list\_matches} & List matches run so far. \\
 & \texttt{get\_match} & Return a previous match's per-seed results and sampled replay without re-running it. \\
\midrule
Analysis & \texttt{analyze\_match} & Run model-written Python over a match's full recording and return only the function's result. The recording exposes per-seed winner, termination reason, step count, positions, velocities, edge distances, contacts and contact forces, tipping, actions and \texttt{qpos}. \\
 & \texttt{diagnose\_physics} & Report a bot's gear, inertia, traction and mass breakdown. \\
 & \texttt{probe\_obs} & Show the observation dictionary the controller actually receives: keys, shapes and sample values. \\
\midrule
Library & \texttt{save\_bot} & Capture the draft as a named, complete version together with its validation and qualification results and the model's design intent (summary, design strategy, hardware plan, combat plan). \\
 & \texttt{list\_bots} & List saved bots with status and summaries. \\
 & \texttt{get\_bot} & Copy a saved bot back into the draft, replacing it, to branch from an earlier design. \\
 & \texttt{finish} & End the run early. Selection still runs. \\
\bottomrule
\end{tabular}
\caption{Design Lab Harness tools, as presented to the model. Verification is
static and cheap; qualification adds simulation against the block;
\texttt{run\_match} verifies but does not require qualification, so two
unqualified bots can fight.}
\label{tab:dlh_tools}
\end{table}

\paragraph{Selection.}
After the last turn, every bot saved with \texttt{save\_bot} enters a round
robin of three seeds in each direction, draws worth half a win, ranked by
Bradley--Terry strength. The winner is the run's artifact; an earlier candidate
may beat the final draft. A manual submission, an unsaved draft, or files
written to an output folder cannot override this selection.

\paragraph{Isolation.}
Each run gets a fresh workspace and a frozen snapshot of the engine and tool
code, mounted in a sandbox with no network, no access to other runs, and no
access to test fixtures or other models' bots. The only supplied opponent is
the stationary block.

\subsection{Material Library}
\label{app:material-library}

\textsc{ArtifactArena} uses a fixed material library to make robot mass and contact properties deterministic across submissions. Each robot geometry specifies a material by name using the MJCF attribute \texttt{material="..."}. During validation, the simulator computes each geometry's mass from its volume and the material density,\[ m_g = \rho_{\mathrm{mat}(g)} V_g , \]
and overwrites the geometry friction using the corresponding material friction coefficient. User-specified \texttt{mass}, \texttt{density}, body mass, and \texttt{inertial} tags are not used for final physics. If a geometry omits a valid material, the verifier assigns the default material \texttt{foam}.

\begin{table}[h]
\centering
\small
\begin{tabular}{llrr}
\toprule
Material & Description & Density ($\mathrm{kg}/\mathrm{m}^3$) & Friction \\
\midrule
\texttt{foam}          & Structural foam          & 200   & 0.50 \\
\texttt{plastic}       & HDPE plastic             & 950   & 0.25 \\
\texttt{rubber}        & High-grip rubber         & 1200  & 1.60 \\
\texttt{carbon\_fiber} & Carbon fiber composite   & 1700  & 0.35 \\
\texttt{aluminum}      & Aluminum alloy           & 2700  & 0.45 \\
\texttt{titanium}      & Titanium alloy           & 4500  & 0.40 \\
\texttt{steel}         & Mild steel               & 7800  & 0.60 \\
\texttt{tungsten}      & Tungsten alloy           & 19300 & 0.40 \\
\bottomrule
\end{tabular}
\caption{Material palette used by the \textsc{ArtifactArena} validation pipeline. The density determines geometry mass from volume, while the friction
coefficient determines the sliding contact friction assigned to robot geometries.}
\label{tab:material-library}
\end{table}
  
\section{Compute}
We use a server with ~446 CPU cores to run our simulation experiments. LLMs are accessed through API calls to 3rd party model providers ~\cite{dspy}.

\section{Model Configurations}
We evaluate 21 LLMs across nine families (Anthropic, OpenAI, Google, xAI, DeepSeek, Moonshot AI, Qwen, and Z.ai), listed in full in Table~\ref{tab:models}. Each model has a single YAML configuration that specifies its API identifier, sampling parameters, and any reasoning or thinking flags. This configuration is loaded at run time and forwarded through LiteLLM to the appropriate provider API. We use DSPy \cite{dspy} for our prompts.

\begin{table*}[t]
  \centering
  \footnotesize
  \caption{Large language models evaluated in the cross-model
  tournament. The Configuration column gives the exact reasoning or
  thinking setting we used; the Max output tokens column gives the
  output budget passed on every call, which is each provider's
  documented maximum for that model.}
  \label{app:model_library}
  \begin{tabular}{@{}llll@{}}
    \toprule
    Family & Model & Configuration & Max output tokens \\
    \midrule
    Anthropic   & Claude Fable 5.1             & Effort = high                        & 128{,}000 \\
    Anthropic   & Claude Fable 5               & Effort = high                        & 128{,}000 \\
    Anthropic   & Claude Sonnet 5              & Effort = high                        & 128{,}000 \\
    Anthropic   & Claude Opus 4.8              & Adaptive thinking, effort = high     & 128{,}000 \\
    \addlinespace[2pt]
    OpenAI      & GPT-6 Astra                  & Reasoning effort = high              & 128{,}000 \\
    OpenAI      & GPT-5.6 Luna                 & Reasoning effort = high              & 128{,}000 \\
    OpenAI      & GPT-5.6 Sol                  & Reasoning effort = high              & 128{,}000 \\
    OpenAI      & GPT-5.6 Terra                & Reasoning effort = high              & 128{,}000 \\
    OpenAI      & GPT-5.5                      & Reasoning effort = high              & 128{,}000 \\
    OpenAI      & GPT-5.4                      & Reasoning effort = high              & 128{,}000 \\
    OpenAI      & GPT-5.3-Codex                & Reasoning effort = high              & 128{,}000 \\
    \addlinespace[2pt]
    Google      & Gemini 3.1 Pro Preview       & Reasoning effort = high              & 65{,}536 \\
    Google      & Gemini 3.8 Flash             & Reasoning effort = high              & 65{,}535 \\
    \addlinespace[2pt]
    xAI         & Grok 4.6                     & Reasoning effort = high              & 131{,}072 \\
    xAI         & Grok 4.5                     & Reasoning effort = high              & 131{,}072 \\
    xAI         & Grok 4.20 0309 Reasoning     & Native reasoning                     & 32{,}000 \\
    \addlinespace[2pt]
    DeepSeek    & DeepSeek V4 Pro (0813)       & Thinking enabled                     & 65{,}536 \\
    \addlinespace[2pt]
    Moonshot AI & Kimi K3                      & Reasoning effort = high              & 1{,}048{,}576 \\
    \addlinespace[2pt]
    Qwen        & Qwen3.8-2.4T-A95B            & Native reasoning                     & 81{,}920 \\
    \addlinespace[2pt]
    Z.ai        & GLM-5.3                      & Reasoning effort = high              & 131{,}072 \\
    Z.ai        & GLM-5.2                      & Reasoning effort = high              & 131{,}072 \\
    \bottomrule
  \end{tabular}
  \label{tab:models}
\end{table*}

\section{Simulation \& Harness Code}

\textsc{ArtifactArena} is built on MuJoCo ~\cite{6907751}, and the same simulator also runs in the browser, where it is used in two ways: 1) A \emph{live match} where the browser does the simulation itself: it composes the arena, steps the physics, calls both controllers every control step, applies the termination rules, and renders the result as it goes. In 2) for a \emph{recorded match}, the browser does not re-simulate but downloads the trajectory stored when the match was played on the cluster, uses the engine's forward kinematics to recover every body pose from it, and renders the game. The user also has full control over the camera and are able to zoom-in/out and pan to watch the match from any angle. In both cases the code doing the work is the same C physics engine and the same Python match code as on the cluster, not a JavaScript re-implementation.

We now describe the engine, the match code, and the \textsc{ArtifactArena} package in detail. 

\paragraph{Physics engine.} We compile the unmodified MuJoCo~3.10.0 source to WebAssembly with Emscripten into a 9MB binary that the html page loads inside a Web Worker.

\paragraph{Match code.} Once we have two verified robots from two models, we compose the two robots into the arena, seeded spawn placement, the settle phase, the control loop, observation construction, and the win, loss and draw rules. We bundle these modules and run them in Pyodide, a CPython build for the browser. A small shim module presents the \texttt{mujoco} Python API on top of the WebAssembly engine, so the harness code imports \texttt{mujoco} as usual and does not know it is in a browser. Submitted controllers are ordinary Python and run unchanged.

\paragraph{Numerical agreement.}
We faced an issue where the browser match and a cluster match outcomes differed between the two builds. Math library functions such as sine and cosine are not correctly rounded in every libm, so the WebAssembly build wraps them with the implementations from the tournament host's glibc. Convex collision detection has a degenerate case for exactly parallel cylinders, common in wheeled robots, where the two builds could pick opposite support points; a small patch breaks the tie deterministically. We then run a native-versus-browser suite over fixture matches: the compiled model (masses, inertias, geometry) must match to $10^{-9}$, the seeded initial state must match bit for bit, settle and driven-probe trajectories must stay within tolerance, repeated browser runs must be bit-identical, and the winner and termination reason must agree. Bit-for-bit equality with native over a full contact-rich match is not achievable across compilers, since differences at the last bit grow chaotically; we therefore state parity as outcome agreement plus bounded trajectory divergence, and the site labels replays as reconstructions.

\paragraph{Exact replays from trajectories.} Our package allows the browser engine to be deterministic, we do not ship video. Each game in the release is stored as its gzipped sequence of generalized coordinates (one vector per control step). To replay, the browser recompiles the composed model, runs MuJoCo's forward kinematics on each stored frame to recover every body pose, and renders the result with three.js.

\paragraph{Playing against the leaderboard.} The same worker can compose a match from a user's \texttt{robot.xml} and \texttt{controller.py} against any released artifact and run it under the tournament rules, so the platform doubles as a development environment for human and human-plus-AI entrants. A 300-second match runs at roughly real time on a laptop.

\paragraph{Human-AI Robot Building.} Having the simulator in the browser also allowed us to create a 3D interactive bot editor. The page recompiles the robot in MuJoCo on every edit and shows the result as a live, physically settled preview rather than a static drawing, with each body, geom and joint in the XML linked to the corresponding part in the 3D view. The user can also step back through their edit history, and download the pair of bot geometry and controller files when ready for submission.

\paragraph{Release.} The harness, the verifier, the WebAssembly build script and patch, the Pyodide bundle, the parity suite and the web viewer will be released with the paper.

\section{Model Prompt}
\label{app:model_prompt}

This is the complete prompt given to each model for zero-shot sampling. The placeholder \texttt{\{mujoco\_version\}} is filled in at run time.

\begin{lstlisting}[language=,breaklines=true,basicstyle=\ttfamily\scriptsize,columns=fullflexible,keepspaces=true]
Design a robot to maximize its win rate in the tournament described below. Provide the robot's hardware as MJCF XML and its controller as Python.

The rules and technical descriptions below define the game and its simulation environment. They do not prescribe what robot to build.

# Competition

Your robot will compete in a round-robin tournament, facing every other robot in a 1v1 match. Each match takes place on an elevated circular platform, 15 m in diameter, with randomized robot starting positions. A robot loses if any part of it touches the floor below the platform. Starting 10 seconds into the round, inactivity is checked every control step. A robot loses if its two furthest center-of-mass positions in the preceding 10 seconds are less than 0.5 m apart, measured in XYZ. You win if your opponent loses before you do. If both robots trigger inactivity loss on the same step, the robot whose center of mass is more than 1 cm higher wins; otherwise, the round is a tie. If your robot causes the physics simulation to become unstable (NaN/Inf in forces), the match terminates and counts as a loss. If neither robot loses before time expires, the match is a draw.

A controller error counts as a loss. Match outcomes are resolved after each control step. If multiple loss conditions occur on the same step, the highest-priority condition determines the result: controller error, floor contact, size violation, inactivity, then physics instability. If both robots lose under that condition, the match is a tie, except for the inactivity height tiebreaker described above. Loss conditions take precedence over time expiration.

## MuJoCo World

The simulation runs in MuJoCo {mujoco_version} with gravity of 9.81 m/s^2. The robot's root body must have a freejoint, allowing translation and rotation in all three dimensions. The controller is called every 0.01 seconds of simulated time, with 40 physics simulation steps between calls. There is no per-call time limit; the simulation waits for the controller to return. Explore any design permitted by the simulation and build rules. Gravity applies fully to every body: any `gravcomp` attribute is removed before validation and reported.

Think from first principles and physics when designing your robot. Each API request has a 100-minute wall-clock limit and a maximum output length; a request that exceeds either ends the run.

## Output format

Return one complete robot design using these six named sections in this order. Give concise explanations, justifications, and calculations in the planning sections.

The XML and Python must be fully implemented, mutually consistent, and self-contained, with no placeholders, ellipses, omitted helpers, or dependencies on external files.

### 1. `name`

A creative, memorable name that hints at the robot's design or strategy.

### 2. `design_strategy`

Briefly describe the robot's overall concept, how it will win, and how its hardware and controller work together.

Explain the key hardware and controller design choices, why you think they will help the robot win, and which aspects carry the greatest risk or uncertainty.

### 3. `hardware_plan`

Describe how the hardware works. Provide calculations evaluating its compliance with the build rules. Account for geometry and motor mass, including total mass, and check the robot's overall dimensions. Explain how you chose actuator strengths and mechanism dimensions based on the forces, torques, and motion required by the design.

End by explicitly mapping each hardware mechanism to the attack, movement, and defense strategies described in `design_strategy`.

### 4. `combat_plan`

Describe how the controller works. What observations does it make use of? How does it use those observations to choose actuator commands and carry out the strategies described in `design_strategy`?

Explain how it handles different starting positions and orientations, opponent behavior, proximity to the platform edge, and recovery from unfavorable situations.

### 5. `robot_xml`

Provide the complete robot MJCF XML in one fenced `xml` code block, ready to save as `robot.xml`.

Follow the `rules` and `mjcf_syntax` sections, including the material palette and the mass, size, and structure constraints.

Every body, geom, internal joint, motor, mesh, site, and tendon definition must include a brief `description="..."` attribute explaining what it does and why it exists in relation to the other parts. The root joint and container tags do not need descriptions. The arena removes these attributes before MuJoCo compilation and uses them to explain and visualize the design.

### 6. `controller_code`

Provide the complete Python controller in one fenced `python` code block, ready to save as `controller.py`.

Implement `def policy_step(obs) -> dict[str, float]`.

Controller state, including globals, persists between `policy_step` calls within a match. Every qualification and tournament match starts with a fresh controller state. This including repeated matches against the same opponent. `random` and `np.random` are private to your controller and re-created for every match; they never affect your opponent. Writing attributes on shared library objects (modules, classes) is a controller error. So is registering a class on a shared abstract base class.

Use the observations and execution environment defined in `obs_schema`. Return a dictionary mapping every actuator name in `robot_xml` to a float in `[-1, 1]`. Actuator names must match the XML exactly. Comment the code to explain its behavior.

Check that every execution path returns valid commands for every actuator.

Before qualification, `policy_step` is called once on a real observation of your robot; an exception or a bad return there is reported as a validation failure.

## Qualification Round

To qualify, your robot must not lose each of three 20-second rounds against a simple baseline block. Your goal is to make a robot that wins all three rounds, but not losing is the minimum bar to qualify. Both the block and your robot start at randomized positions. The block has no motors and can freely translate, rotate, and topple. It is exempt from inactivity loss. If your robot fails to qualify, it forfeits and loses the tournament.

The block is a 1.2 x 1.2 x 0.25 m plastic slab of about 342 kg with the palette's plastic friction.

## Tournament Phase

Your robot faces every other robot in independently scored rounds lasting up to 300 seconds. Tournament standings use Bradley-Terry ratings, with draws counted as half a win for each robot. Both robots start at randomized positions in each round.

## Rules

```yaml
# Arena configuration
arena:
  shape: circle
  radius_m: 7.5  # diameter 15m
  thickness_z_m: 2.0

# Robot constraints
robot:
  # Mass constraints
  # Geometry mass: volume x material density for each geom.
  # Motor mass: sum(priced gear for each motor) x motor_mass_per_gear. The priced gear is the gear magnitude; for a motor on a fixed tendon it is the gear magnitude x sum(|coef|) over that tendon's joints.
  # Gear magnitude is abs(gear) for a scalar, or sqrt(sum(component^2)) for a vector.
  # Total robot mass includes both geometry and motor mass.
  mass:
    min_total_kg: 25.0
    max_total_kg: 800.0

  # Control constraints
  control:
    motor_mass_per_gear: 0.01
    max_dof_damping: 600.0  # Internal joint/tendon linear damping: 0-600, no additional mass cost.
    max_frictionloss: 600.0  # Per internal joint DOF or tendon; no additional mass cost.
    # Root-joint damping, frictionloss, and stiffness are fixed at 0.
    # Configurable environment setting: 0 disables gear clamping.
    # When enabled, max gear = body_inertia_min x gear_clamp_ratio.
    gear_clamp_ratio: 0

  # Structure limits
  structure:
    max_bodies: 2000
    max_geoms: 2000
    max_actuators: 2000
    min_actuators: 1

  # Maximum axis-aligned bounding box dimensions, in metres
  size:
    max_x_span_m: 2.44
    max_y_span_m: 2.44
    max_z_span_m: 3.05
```

Size limits apply to the robot's initial pose defined by its XML, measured along the world X, Y, and Z axes before spawn rotation and settling. During a match the limits are also checked on every control step in your robot's own root-body frame: if your robot's extents along its root axes exceed 2.44 x 2.44 x 3.05 m at any step, you lose the round. Validation measures the initial pose in the root-body frame as well and rejects a robot outside the box. Design mechanisms to stay inside the box while moving.

The arena fixes `condim="6"`, enabling sliding, torsional, and rolling friction. It sets sliding friction from each geom's material, torsional friction to `0.005`, and rolling friction to `0.0001`, overriding model-specified values.

Collision settings are controlled by the simulation environment. Do not specify geom `contype`, `conaffinity`, `condim`, `friction`, `priority`, `solmix`, `solref`, `solimp`, `margin`, or `gap`, including through `<default>` blocks. Do not define `<contact>` blocks, contact pairs, or collision exclusions. The validator reports and ignores these overrides. Robot geoms use `contype="1"`, `conaffinity="1"`, `priority="2"`, `solmix="1"`, `solref="0.02 1"`, `solimp="0.9 0.95 0.001 0.5 2"`, and `margin="0" gap="0"`.

The platform's own friction never applies to a robot-platform contact: robot geoms have contact priority 2 and the platform priority 1, so your material's friction is your traction. Priority, condim and the other contact attributes are set by the environment and cannot be changed by your XML.

## MJCF syntax

MJCF is MuJoCo's XML format. You define a tree of physical parts, connect them
with joints, then attach motors to drive those joints.

### COORDINATE SYSTEM

- Robot-local axes: +X forward, +Y left, +Z up.
- World axes: XY is horizontal; +Z is up (gravity acts along -Z).
- Units: meters, kilograms, seconds, radians.

### DOCUMENT STRUCTURE

Your MJCF XML must use the following structure:

```xml
<mujoco>
  <asset>
    <!-- inline mesh definitions, if needed -->
  </asset>
  <worldbody>
    <!-- your robot's root body goes here -->
  </worldbody>
  <tendon>
    <!-- optional tendon definitions -->
  </tendon>
  <actuator>
    <!-- motors that drive joints or tendons go here -->
  </actuator>
</mujoco>
```

Do not include an `<option>` block: global physics settings (timestep, gravity, integrator, solver) belong to the arena; any `<option>` in your XML is removed before validation and reported.

The `<asset>` block may contain inline mesh definitions for SDF geoms. Include it only if your robot uses inline meshes for SDF geoms. Any other child of `<asset>` (materials, textures) is removed before validation and reported.

The arena sets compiler `angle="radian"` when validating and composing the match. Any other `<compiler>` attribute in your XML is removed before validation and reported.
Optional `<default>` blocks may define shared joint, tendon, and motor settings. Do not use `<frame>`, `<replicate>`, `<attach>`, `<composite>`, `<flexcomp>`, or `<include>` elements; the validator rejects them. Set `pos` and orientation on bodies and geoms directly.

### BUILDING BLOCKS

#### 1. BODY

An MJCF `<body>` tag defines a rigid body that may contain geoms, joints, and child bodies.

Place exactly one root `<body>` directly inside `<worldbody>` and nest every other robot body beneath it.

- `pos="x y z"` sets the body's origin in its parent's coordinate frame.
- Bodies nest: a child body inherits its parent's position and orientation. A joint allows relative motion; without a joint, the bodies are rigidly connected.
- Each body with a joint or freejoint must have a geom in itself or a rigidly attached descendant. Descendants reached through another joint or freejoint do not count.
- The root body MUST have a `<freejoint/>` so the robot can move freely in 3D.
- `description="..."` -- REQUIRED. A short text explaining what this body does
  and WHY it exists (e.g. "Low steel ballast to lower center of gravity").
  Stripped before MuJoCo compilation; used by the app to display design intent.

```xml
<body name="root_body" pos="0 0 0.15" description="Root coordinate frame for the robot">
  <freejoint name="root"/>
  <!-- geoms and child bodies go here -->
</body>
```

#### 2. GEOM

An MJCF `<geom>` tag defines a geometric shape rigidly attached to its containing `<body>`, used for visualization, collision, and calculating mass and inertia.

```xml
<geom name="geom_name" type="geom_type" material="material_name" description="purpose"/>
```

- Supported geom types: `box`, `cylinder`, `sphere`, `capsule`, `ellipsoid`, and `sdf`. `type="mesh"` is not accepted; any mesh can be expressed as an `sdf` geom from the same vertices and faces.
- `material="name"` selects a material from the palette below, determining density and sliding friction. Geom mass is volume x material density.

- Every geom must have a unique name, set by its `name` attribute. A geom without a name fails validation.

For `box`, `cylinder`, `sphere`, `capsule`, and `ellipsoid` geoms, `size` values are radii or half-lengths. For example, a box with `size="1 1 1"` measures 2 x 2 x 2 m.

- box: `size="half_x half_y half_z"`
- cylinder: `size="radius half_height"`
- sphere: `size="radius"`
- capsule: `size="radius half_length"`
- ellipsoid: `size="radius_x radius_y radius_z"`

An SDF (signed distance field) gives the distance to a shape's surface: negative inside, zero on the surface, and positive outside. An SDF geom can be created by defining a `<mesh>` in `<asset>` using vertices and triangular faces, then referencing it with `<geom type="sdf" mesh="mesh_name"/>`. MuJoCo computes the SDF from the mesh and uses it for collision detection, preserving concave features.

A mesh defines a surface using vertices connected into triangular faces. `vertex` lists the points, and `face` specifies which groups of three points form triangles. Define its `vertex` attribute as XYZ coordinates in meters and its `face` attribute as triples of zero-based vertex indices. The surface must be closed and outward-facing (every edge shared by exactly two triangles, positive enclosed volume). The simulation environment computes mass and inertia from the mesh's enclosed volume and material density.

SDF dimensions come from mesh vertices, multiplied along X, Y, and Z by the `<mesh>` attribute `scale="scale_x scale_y scale_z"` (default: `"1 1 1"`). Do not use geom `size` to resize an SDF. The geom's `pos` and orientation attributes set its position and orientation relative to its body.

A mesh asset adds no mass itself. Each geom that uses it has mass equal to the mesh's enclosed volume x the geom's material density.

Define mesh vertices and faces directly in `robot.xml`; external mesh files and custom SDF functions are not supported.

MuJoCo {mujoco_version} assigns mesh-derived SDF sample signs using the normal of a nearest triangle. Near some edges and corners, this can incorrectly classify empty space as solid and produce contacts outside the intended surface. Rounding affected edges and corners with multiple facets so normals change gradually is one possible mitigation, not a requirement or a guarantee of correct collisions. Subdividing an unchanged flat surface does not change its sharp boundaries.

SDF syntax example: Define the mesh in `<asset>` directly under `<mujoco>`:

```xml
<asset>
  <mesh name="surface_a"
        vertex="0 0 0  0.6 0 0  0 0.4 0  0 0 0.3"
        face="0 2 1  0 1 3  0 3 2  1 2 3"
        description="Closed surface with explicit vertices and outward-facing triangles"/>
</asset>
```

Reference the mesh from an SDF geom inside a `<body>`:

```xml
<geom name="surface_geom" type="sdf" mesh="surface_a" material="aluminum"
      description="Uses the inline surface for SDF collision geometry"/>
```

**Orientation:**

A geom's orientation specifies its rotation relative to its immediate parent body. By default, the geom's axes align with that body's axes. For cylinders and capsules, the geom's local Z-axis passes through the centers of both ends. Set orientation on `<geom>` using one of the following two options:

**1. Euler angles (`euler`)**

`euler="rx ry rz"` specifies rotations in radians about the geom's local X, then Y, then Z axes. Each rotation uses the axes resulting from the preceding rotations. Positive rotations follow the right-hand rule.

Euler examples:

`euler="1.5708 0 0"` rotates the geom approximately 90 deg about X, pointing its local +Z axis along its parent body's -Y axis.

`euler="0 1.5708 0"` rotates the geom approximately 90 deg about Y, pointing its local +Z axis along its parent body's +X axis.

**2. Axis directions (`xyaxes`)**

`xyaxes="x1 y1 z1 x2 y2 z2"` specifies the geom's X-axis direction followed by its Y-axis direction, expressed in its immediate parent body's coordinates. MuJoCo makes these axes perpendicular and normalizes them, then computes Z = X x Y.

Geom orientation and joint axis are independent. Choose their relationship to produce the intended motion. For example, a cylinder intended to rotate about its symmetry axis must have that axis aligned with its hinge axis.

#### 3. JOINT

Defines how a child body can move relative to its parent.

- Without a joint, a child body is rigidly welded to its parent.
- Place the joint inside the child body (not the parent).

Geoms on independently moving branches of your robot can collide with each other. Check that moving parts can complete their intended motion without unintended collisions.

For hinge and slide joints, `axis="x y z"` sets the rotation or translation direction in the child body's coordinates. For hinges, the rotation axis passes through `pos="x y z"` in those same coordinates. Defaults are `axis="0 0 1"` and `pos="0 0 0"`.

To limit a hinge or slide joint's movement, set `limited="true"` and `range="min max"`. Hinge limits are angles in radians; slide limits are distances in meters. Set `limited="false"` for unrestricted movement.

Hinge and slide joints start at `ref` (default: `0`), corresponding to the pose defined by the XML. Before the first controller call, the environment simulates 50 physics steps (0.0125 seconds) with zero motor commands, then sets all joint velocities to zero. Joint positions may change during settling. Observations provide current joint positions and velocities in `my_robot["joints"]`.

**Joint types:**

**hinge:**  Rotates a child body around the specified local axis. `axis="0 1 0"` means rotate around the local Y axis.

```xml
<joint name="rotation_joint" type="hinge" axis="0 1 0"/>
```

**slide:**  Translates a child body along the specified axis. `axis="1 0 0"` means slide along the X axis.

```xml
<joint name="translation_joint" type="slide" axis="1 0 0"/>
```

**ball:**   Allows rotation in any direction around a shared connection point (3 rotational DOF), with no relative translation at that point. `pos="x y z"` locates the joint in the child body's frame. `axis=` is ignored because there is no single rotation axis. MuJoCo represents its orientation as a unit quaternion (w, x, y, z).

```xml
<joint name="spherical_joint" type="ball" pos="0 0 0"
       description="Allows three-axis rotation at the connection"/>
```

A body with a ball joint cannot also have a hinge or another ball joint. It may have slide joints. The joint can be passive. A single motor provides one commanded torque component, not independent control of all three rotations. The joint's observation provides its orientation quaternion and full local angular-velocity vector.

**free:**   6-DOF (translate + rotate). Only for the root body of the robot.

```xml
<freejoint name="root"/>
```

Internal joints may use scalar `damping="d"` from 0 to 600 (default: 0), with no additional mass cost. Damping applies F = -d x velocity to sliding joints (d in N.s/m) and torque = -d x angular_velocity to rotating joints (d in N.m.s/rad). The environment sets root-joint damping, frictionloss, and stiffness to 0, regardless of your XML.

Internal joints and tendons may also use `frictionloss="f"` from 0 to 600 (default: 0), with no additional mass cost. Friction loss provides passive resistance to motion, including resistance to starting from rest, and may be combined with damping. Values are in N for slide joints and spatial tendons, and N.m per rotational degree of freedom for hinge and ball joints. For fixed tendons, the limit applies to the tendon-coordinate force; each joint's resistance is multiplied by its tendon coefficient. Values must be finite.

Internal joint and tendon springs may use scalar `stiffness="k"` >= 0, with no upper cap or additional mass cost. Stiffness must be finite. Springs must start unloaded in the XML's initial pose, before settling; motion may then load them. For hinge/slide springs, set springref equal to ref (both default to 0). Restoring force/torque is -k x displacement (k in N/m or N.m/rad). Tendon springlength sets a rest length or a two-value unloaded interval; the initial length must lie within it. If omitted, MuJoCo computes it from the joint spring-reference pose, so match every joint's springref to ref. The environment fixes joint, tendon, and actuator armature at 0.

#### 4. MOTOR (actuator)

An MJCF `<motor>` applies force or torque through a transmission. Define motors inside `<actuator>`.

- Only `<motor>` actuators are allowed.
- Every `<motor>` must have a unique `name`: your controller returns a dictionary keyed by motor name, so an unnamed motor fails validation.
- Set `joint="joint_name"`, `jointinparent="joint_name"`, or `tendon="tendon_name"` to select what the motor drives. These drive an internal hinge, slide, or ball joint, or a tendon.
- Motor forces and torques must act between parts of your robot. Do not actuate the root freejoint, use a site motor without `refsite`, or anchor a transmission to the world or another robot.
- Motor force or torque equals `gear x motor command`. For hinge joints, torque is in N.m and positive torque follows the right-hand rule around the joint's `axis`. For slide joints, force is in N and positive force acts along the joint's `axis`.
- For ball joints, set `gear="gx gy gz"`. These values multiplied by the motor command give the torque vector in N.m in the child body's coordinates for `joint`, or the parent body's coordinates for `jointinparent`; positive components follow the right-hand rule. A single gear value specifies only the X component; the others default to 0.

- Relative-site motors use `site="site_name" refsite="reference_site_name"`. Both sites must belong to your robot; either may be on the root body. `gear="fx fy fz tx ty tz"` selects translational and rotational components in the reference site's frame, scaled by the motor command. Translation components correspond to forces in N; rotation components correspond to torques in N.m.
- Slider-crank motors use `cranksite="crank_site_name" slidersite="slider_site_name" cranklength="length"`. Both sites must belong to your robot. The connecting-rod length is in meters, and the slider moves along the slider site's local Z axis. Scalar `gear x motor command` applies force in N to the slider-displacement coordinate. MuJoCo models the transmission; represent its physical parts with material geoms.

- Motor commands range from -1 to 1.
- Higher absolute gear provides greater force or torque and adds motor mass. For a motor on a fixed tendon, the priced gear is |gear| multiplied by the sum of the absolute joint coefficients of that tendon. A motor on a spatial tendon is priced at |gear|, the force it applies along the tendon.
- Motor mass is added to the first geom of its mounting body: the joint body for joint motors, the `site` body for relative-site motors, the `slidersite` body for slider-crank motors, and the root body for tendon motors. If that body has no geom, the environment searches its rigidly attached descendants depth-first in XML order, skipping bodies with joints or freejoints. The added mass follows that geom's shape and position, contributing to the robot's simulated mass, center of mass, and inertia.

- Motors have no arena-imposed speed, power, energy, or thermal limits. Their available force or torque does not decrease with speed.

```xml
<actuator>
  <motor name="actuator_a" joint="rotation_joint" gear="1250"/>
</actuator>
```

#### 5. TENDON

An MJCF tendon transmits force between parts. Define tendons inside an optional `<tendon>` block directly under `<mujoco>`.

- A `<spatial>` tendon represents a cable routed through sites on robot bodies. A site is a massless reference point, defined as `<site name="site_name" pos="x y z"/>` inside a body.

Spatial tendons can wrap around sphere or cylinder geoms.

- `<fixed>` defines length = sum(coef x joint_position) for hinge/slide joints.
- Tendons may have springs, scalar damping from 0 to 600 (force = -d x length_rate), and length limits (`limited="true"` `range="min max"`). Spatial lengths are in meters; fixed tendon units follow their joint coordinates and coefficients.

- Tendons must remain entirely within the robot; world anchors and connections to other robots are forbidden. `<equality>` constraints are not allowed; the validator rejects them. `<pulley>` elements are not allowed either.
- Tendons and sites add no mass or collision geometry. Represent physical hardware with material geoms. Include descriptions on sites and tendons.

- A motor may use `tendon="name"` instead of `joint="name"`. It applies gear x motor command to the tendon-length coordinate; negative force pulls a spatial tendon shorter.

```xml
<tendon>
  <spatial name="cable" stiffness="1000" damping="20"
           description="Elastic connection between the chassis and arm">
    <site site="chassis_anchor"/>
    <site site="arm_anchor"/>
  </spatial>
  <fixed name="coupling" description="Couples two joint coordinates">
    <joint joint="joint_a" coef="1"/>
    <joint joint="joint_b" coef="-1"/>
  </fixed>
</tendon>
<actuator>
  <motor name="winch" tendon="cable" gear="2000"
         description="Pulls the cable to move the connected mechanism"/>
</actuator>
```

### HOW TRANSFORMS WORK (critical for correct assembly)

Every body's `pos=` is RELATIVE TO ITS PARENT, not the world. When a parent body moves, ALL its children move with it automatically.

Example: parent body at `pos="0 0 0.15"`, child body inside it at `pos="0.2 0.25 -0.05"` -> child's world position = (0+0.2, 0+0.25, 0.15-0.05) = (0.2, 0.25, 0.10) -> when the parent moves to x=1, the child moves to (1.2, 0.25, 0.10) automatically

Geom `pos=` is relative to its OWN body (usually "0 0 0" or omitted). Joint in a child body only moves THAT child, not siblings.

### MATERIAL PALETTE (part of mjcf_syntax)

Pick a material for every geom. Geometry mass is volume times material density; for a mesh-derived SDF, use the enclosed mesh volume. Do not use `mass=`, `density=`, body `mass=`, or `<inertial>`. An optional `<asset>` block may contain inline `<mesh>` definitions referenced by SDF geoms. Preserve these mesh definitions when processing the submission. Palette materials are injected by the arena; do not define or override materials, textures, or other assets. External asset files and custom plugin declarations are not part of this submission format.

```yaml
# Material Palette
#
# Each material defines physics (density, friction) and visual (rgba, specular,
# shininess) properties. Pick a material per geom (e.g., material="steel").
# Geom mass is volume x material density. Sliding friction comes from the material's friction coefficient.
#
# Visual properties are injected as MuJoCo <material> assets so material="steel"
# is both a physics lookup AND a valid visual reference.

# Default material auto-assigned to geoms missing material= attribute.
# Must be a key in the materials dict below.
default_material: foam

materials:
  foam:
    density_kg_m3: 200
    friction: 0.5
    desc: "Structural foam"
    rgba: "0.92 0.90 0.85 1"
    specular: 0.0
    shininess: 0.0
  plastic:
    density_kg_m3: 950
    friction: 0.25
    desc: "HDPE plastic"
    rgba: "0.95 0.95 0.95 1"
    specular: 0.1
    shininess: 0.1
  rubber:
    density_kg_m3: 1200
    friction: 1.6
    desc: "High-grip rubber"
    rgba: "0.12 0.12 0.12 1"
    specular: 0.0
    shininess: 0.0
  carbon_fiber:
    density_kg_m3: 1700
    friction: 0.35
    desc: "Carbon fiber composite"
    rgba: "0.10 0.10 0.12 1"
    specular: 0.3
    shininess: 0.2
  aluminum:
    density_kg_m3: 2700
    friction: 0.45
    desc: "Aluminum alloy"
    rgba: "0.75 0.75 0.78 1"
    specular: 0.5
    shininess: 0.3
  titanium:
    density_kg_m3: 4500
    friction: 0.4
    desc: "Titanium alloy"
    rgba: "0.60 0.58 0.55 1"
    specular: 0.4
    shininess: 0.25
  steel:
    density_kg_m3: 7800
    friction: 0.6
    desc: "Mild steel"
    rgba: "0.55 0.55 0.60 1"
    specular: 0.6
    shininess: 0.5
  tungsten:
    density_kg_m3: 19300
    friction: 0.4
    desc: "Tungsten alloy"
    rgba: "0.35 0.35 0.38 1"
    specular: 0.4
    shininess: 0.3
```

## obs_schema

`policy_step(obs)` receives a Python dictionary whose keys are strings. Every name listed under `fields` below is an exact top-level key in `obs`, accessed as `obs["key_name"]`. All listed keys are present on every controller call; lists and dictionaries may be empty where specified.

The YAML below documents each value's type, shape, units, and meaning; the values received by the controller are Python objects, not the descriptive strings shown here. `np.ndarray[3]` means a NumPy array of shape `(3,)`, `dict[str, float]` means a dictionary mapping string keys to floats, and `list[dict]` means a list of dictionary records. Nested record keys are specified in "Detailed physical-state records" below. The labels `fields` and `controller_environment` organize this specification; neither is a key in `obs`.

```yaml
fields:
  # === POSITION ===
  my_pos: "np.ndarray[3] - entire robot's center of mass [x, y, z] in world coordinates (meters); XY is the horizontal plane, +Z is up"
  opponent_pos: "np.ndarray[3] - opponent's center of mass, using the same convention as my_pos"

  # === ORIENTATION (scalars, radians) ===
  my_yaw: "float - root-body heading in radians, from -pi to pi (0 = world +X, pi/2 = world +Y)"
  my_pitch: "float - root-body pitch in radians, from -pi/2 to pi/2 (0 = level, positive = nose down)"
  my_roll: "float - root-body roll in radians, from -pi to pi (0 = level, positive = left side up, right side down)"
  opponent_yaw: "float - opponent's root-body heading, using the same convention as my_yaw"
  opponent_pitch: "float - opponent's root-body pitch, using the same convention as my_pitch"
  opponent_roll: "float - opponent's root-body roll, using the same convention as my_roll"

  # === VELOCITY ===
  my_velocity: "np.ndarray[3] - velocity of the root-body origin [vx, vy, vz] in world coordinates (m/s)"
  my_angular_velocity: "np.ndarray[3] - angular velocity [wx, wy, wz] about the root body's local X, Y, Z axes (rad/s), positive by the right-hand rule"
  opponent_velocity: "np.ndarray[3] - velocity of the opponent's root-body origin, using the same convention as my_velocity"
  opponent_angular_velocity: "np.ndarray[3] - angular velocity about the opponent's root-body local axes, using the same convention as my_angular_velocity"
  my_actuator_velocity: |
    dict[str, float] - maps each motor's XML name to the velocity defined below for its transmission type:
    - Hinge: angular velocity about the joint axis (rad/s), positive by the right-hand rule.
    - Slide: linear velocity along the joint axis (m/s), positive in the axis direction.
    - Spatial tendon: length change per second (m/s), positive when lengthening.
    - Fixed tendon: sum(coef x joint velocity), using signed joint velocities and coefficients.
    - Ball: joint angular velocity about the motor's torque axis (rad/s), positive in the torque direction produced by a positive motor command. Gear magnitude does not scale this value; zero gear returns 0.
    - Slider-crank: slider-displacement rate in m/s, before gear scaling (0 for zero gear).
    - Relative-site: MuJoCo's scalar actuator velocity, including all gear components. It combines the selected relative translation and rotation rates; it is not an individual joint velocity.
  opponent_actuator_velocity: "dict[str, float] - opponent motor velocities, keyed by their XML names, using the same convention as my_actuator_velocity"

  # === DISTANCES ===
  distance_to_opponent: "float - Euclidean distance in XYZ between the robots' centers of mass, my_pos and opponent_pos (meters)"
  my_edge_distance: "float - signed XY distance from your robot's center of mass to the platform edge (meters); positive inside the platform boundary, zero on it, negative outside"
  opponent_edge_distance: "float - signed XY distance from the opponent's center of mass to the platform edge, using the same convention as my_edge_distance"

  # === CONTACT ===
  opponent_contact: "bool - True if any of your robot's geoms contact an opponent geom"
  opponent_contact_force: "float - sum of force magnitudes across all contacts with the opponent (N); 0 if no contact"
  ground_contact: "bool - True if any of your robot's geoms contact the platform, including its sides"

  # === STABILITY ===
  is_tipping: "float - angle in radians between the root body's +Z axis and world +Z, divided by pi/2 and capped at 1 (0 = upright, 1 = tilted 90 deg or more)"

  # === INACTIVITY ===
  my_inactivity_timer: |
    float - Duration in seconds of the longest continuous period ending now
    during which every pair of recorded center-of-mass positions is less
    than 0.5 m apart in XYZ. Starts at 0 when the round begins.
    Inactivity loss triggers at 10 seconds.
  opponent_inactivity_timer: "float - measured as described for my_inactivity_timer, but for the opponent robot. Always 0 for the qualification block."

  # === TIME ===
  t: "int - completed control steps in this match; starts at 0 on the first policy_step call of each match and increases by 1 per call. Elapsed simulated time = t x 0.01 seconds"
  max_t: |
    int - control-step limit for this match: 2000 for qualification (20 seconds), 30000 for the tournament (300 seconds).
    On every policy_step call, 0 <= t < max_t.
    Scheduled time remaining = (max_t - t) x 0.01 seconds.

  # === ROBOT PROPERTIES (static per match) ===
  my_bounding_radius: "float - approximate robot radius (meters), computed once at initialization as max(XY distance from COM to each geom's center + that geom's bounding-sphere radius). Does not update as the robot moves or articulates."
  opponent_bounding_radius: "float - computed as described for my_bounding_radius, but for the opponent robot"
  ring_radius: "float - arena radius in meters (distance from center to edge)"

  # === SPATIAL GRIDS (2D top-down maps, rebuilt every timestep) ===
  arena_grid: "np.ndarray[N,N] int8 - occupancy map of circles centered on robot COMs using the static bounding radii, not actual geometry. -1=outside ring, 0=free, 1=opponent, 2=self; self overwrites opponent where circles overlap. center cell = ring center. world_x = (col - N//2) * 0.25, world_y = (N//2 - row) * 0.25"
  arena_mass_grid: "np.ndarray[N,N] float32 - uses the same circular masks as arena_grid. -1=outside, 0=free; each occupied cell repeats that robot's entire mass (kg), not mass within that cell. Self overwrites opponent on overlap."
  edge_distance_grid: "np.ndarray[N,N] float32 - signed distance to ring edge per cell. Positive=inside, negative=outside. Static per match."

  # === GAME CONTEXT ===
  game: "dict - reserved for game-specific context; empty ({}) in every match. Your controller may ignore it."

  # === HISTORY (rolling queues, most recent last) ===
  obs_history: "list[dict] - last N observations (N=max_obs_lookback, default 20). Each dict has: my_pos, opponent_pos, my_yaw, my_velocity, opponent_velocity, distance_to_opponent, my_edge_distance, opponent_edge_distance, t, my_actuator_velocity, game. Empty at t=0."
  action_history: "list[dict] - last N actions YOU took (N=max_action_lookback, default 20). Each dict maps actuator_name -> float value. Commands are recorded before motor-specific ctrl_range limits are applied. Empty at t=0."


  # === DETAILED PHYSICAL STATE ===
  control_dt: "float - simulated seconds per controller call (0.01)"
  elapsed_time: "float - elapsed match time in simulated seconds, equal to t x control_dt"
  time_remaining: "float - scheduled time remaining in simulated seconds, equal to (max_t - t) x control_dt"
  platform: |
    dict with the following keys; all distances are in meters:
    - shape: str, "cylinder".
    - center: np.ndarray[3], center of the platform's top surface in world XYZ coordinates.
    - radius: float, platform radius.
    - half_extents: np.ndarray[2], [radius, radius] along world X and Y.
    - top_z: float, platform-top height in world coordinates; equals center[2].
    - floor_z: float, lower-floor height in world coordinates.
  my_mass: "float - total simulated robot mass, including motor mass (kg)"
  opponent_mass: "float - opponent's total simulated mass (kg)"
  my_com_velocity: "np.ndarray[3] - velocity of your entire robot's center of mass in world XYZ (m/s)"
  opponent_com_velocity: "np.ndarray[3] - opponent's center-of-mass velocity in world XYZ (m/s)"
  my_robot: "dict - named geometry, bodies, joints, tendons, motors, and sites, with definitions and current state as specified below"
  opponent_robot: "dict - the same physical information for the opponent; names belong to its XML"
  opponent_surface_distance: "float - minimum separation between the robots' collision shapes (meters), computed numerically; 0 when intersecting"
  opponent_proximity: "list[dict] - closest geom pairs with separation <= proximity_cutoff; records specified below"
  proximity_cutoff: "float - maximum separation included in opponent_proximity (2 meters)"
  proximity_limit: "int - maximum number of geom pairs returned in opponent_proximity (32)"
  proximity_truncated: "bool - True if more than proximity_limit geom pairs were within proximity_cutoff"
  contacts: "list[dict] - current solver contacts involving your robot, with positions and force/torque vectors as specified below"
  contact_impulses: "list[dict] - impulses accumulated over all physics steps since the previous controller call, grouped by geom pair"
  contact_interval: "float - simulated duration covered by contact_impulses (seconds); 0 on the first call, normally control_dt thereafter"

controller_environment: |
  Available libraries (pre-loaded, no other imports allowed):
    import math         # standard math functions (atan2, sqrt, pi, sin, cos, etc.)
    import numpy as np  # also available as bare 'np' (pre-injected into namespace)
    import typing       # type hints only
    import random       # standard PRNG (seed it yourself if you need determinism)
    import collections  # deque, defaultdict, Counter
  All Python builtins are available (`range`, `len`, `min`, `max`, `abs`, `dict`, `list`, etc.), except `open`, `exec`, `eval`, `compile`, `breakpoint`, `exit`, `quit`, `input`, `globals`, `locals`, `vars`, `getattr`, `setattr`, and `delattr`, which are forbidden. File and process access, including NumPy's save/load family, is forbidden.
  Forbidden: os, subprocess, socket, open(), exec(), eval(), pickle, threading
```

### Detailed physical-state records

Body, geom, and site positions and velocities use world XYZ coordinates unless marked `local`. Positions are in meters, linear velocities in m/s, and angular velocities in rad/s. Joint coordinates and frames are specified separately below. Position, velocity, force, torque, impulse, and normal vectors are `np.ndarray[3]`. Quaternions are `np.ndarray[4]` in `[w, x, y, z]` order. In body, geom, and site records, `quaternion` rotates that element's local coordinates into world coordinates. In geom and site records, `local_quaternion` rotates the element's local coordinates into its containing body's coordinates. Joint and inertia quaternion frames are specified below. Values are detached snapshots: retaining or modifying them does not alter the simulation or future observations. Definitions remain constant within a match; positions and motion update on every call. The definitions describe the compiled robot after environment processing, including injected motor mass.

`my_robot` and `opponent_robot` each contain `mass`, `com_position`, `com_velocity`, `root_body` (the root body's name), and six dictionaries keyed by XML names: `bodies`, `geoms`, `joints`, `tendons`, `motors`, and `sites`. Runtime robot prefixes are removed. Unnamed elements receive stable generated names. Look up a root body's full pose and motion in `robot["bodies"][robot["root_body"]]`. Against the baseline block, `obs["opponent_robot"]["root_body"]` is `"block"`, and its body record is `obs["opponent_robot"]["bodies"]["block"]`.

- **Bodies:** `parent` (body name, or `None` for the root), `mass` (kg), `position`, `quaternion`, `linear_velocity` measured at the body origin, `angular_velocity`, `com_position`, and `com_velocity`. Mass and COM refer to this body's own mass, excluding descendants. `local_com` is its COM in body coordinates. `inertia_diagonal` (`np.ndarray[3]`) gives the three principal moments about its COM (kg.m^2); `inertia_quaternion` (`np.ndarray[4]`) orients those principal axes relative to the body.
- **Geoms:** `body` (containing body name), `type`, `size` (`np.ndarray[3]`, MuJoCo radius/half-length conventions, padded to three values), `local_position`, `local_quaternion`, world `position`, `quaternion`, `linear_velocity` at the geom origin, `angular_velocity`, and `friction` (`np.ndarray[3]`, sliding, torsional, and rolling coefficients). SDF geoms have `type="sdf"`, `vertices` (`np.ndarray[N,3]`, XYZ coordinates in meters in the compiled geom's local frame), and `faces` (`np.ndarray[M,3]`, triples of zero-based vertex indices defining triangles). Mesh dimensions come from these vertices; do not apply geom `size` to them. Geom mass is represented in its body's mass and inertia.
- **Joints:** all internal joints, including passive ones. Each record contains `body`, `type`, `axis` and `anchor` in body coordinates, `limited`, `range`, `reference`, `position`, `velocity`, `stiffness`, `damping`, and `frictionloss`. Hinge positions/rates use radians and rad/s, positive by the right-hand rule about `axis`; slide positions/rates use meters and m/s, positive along `axis`. `position` is the current joint coordinate, including its reference offset; `reference` contains its initial XML coordinate. For hinge and slide joints, `position` and `velocity` are floats; `reference`, `damping`, and `frictionloss` are NumPy arrays of shape `(1,)`. For ball joints, `position` and `reference` are quaternion arrays of shape `(4,)` describing joint rotation relative to the initial XML configuration, not world orientation; `velocity`, `damping`, and `frictionloss` have shape `(3,)`. Ball-joint angular velocity is expressed in the child body's local frame. For all joints, `axis` and `anchor` have shape `(3,)`, and `range` has shape `(2,)`. `range` only constrains limited joints; a ball joint's range describes rotation angle, not quaternion components. Root free-joint state is represented by the root body's pose and motion.
- **Tendons:** `type` (`fixed` or `spatial`), `path`, `length` (`float`), `velocity` (`float`), `stiffness`, `damping`, `frictionloss`, `springlength` (`np.ndarray[2]`, unloaded interval), `limited` (`bool`), and `range` (`np.ndarray[2]`). Spatial length/rate use meters and m/s; fixed length/rate are sums of signed `coef x joint position/velocity`. `path` is an ordered list of records: `{"type": "joint", "joint": name, "coef": value}`, `{"type": "site", "site": name}`, `{"type": "sphere" or "cylinder", "geom": name, "sidesite": name or None}`, or `{"type": "pulley", "divisor": value}`.
- **Motors:** `transmission` (`joint`, `jointinparent`, `tendon`, `site`, or `slidercrank`), `target` (the joint, tendon, site, or cranksite name), `gear` (`np.ndarray[6]`, six MuJoCo gear components), `ctrl_limited` (`bool`), and `ctrl_range` (`np.ndarray[2]`). Relative-site motors also provide `refsite`; slider-crank motors provide `slidersite` and `cranklength` (meters). These definitions connect motor commands to the physical mechanism. Commands must still be in `[-1, 1]`; an enabled `ctrl_range` further limits the command used by that motor.
- **Sites:** `body`, `local_position`, `local_quaternion`, world `position` and `quaternion`, `linear_velocity` at the site origin, and `angular_velocity`. Sites identify tendon and motor reference frames.

`opponent_proximity` records contain `my_geom`, `opponent_geom`, `distance`, `my_point`, `opponent_point`, and `closing_speed`. Points are world XYZ. Positive closing speed means the two points' instantaneous velocities are approaching along their separation direction (m/s); it is 0 when separation is effectively zero. It is not a prediction of future motion. Pairs are sorted by separation, with ties resolved by compiled geom order, then limited to 32. `opponent_surface_distance` considers all geom pairs, including those outside the proximity cutoff. Primitive proximity uses convex surface-distance queries. For SDF geoms, proximity uses the original mesh surface, which may differ from MuJoCo's approximate collision surface. A positive proximity distance does not guarantee absence of contact. `contacts` reports the physics solver's contacts. Results are numerical approximations; intersecting-shape witness points need not coincide or define a unique contact location.

Each `contacts` record contains `my_geom`, `other_kind` (`self`, `opponent`, `platform`, `floor`, or `environment`), `other_geom`, `position`, `normal_toward_me`, `force_on_me` (N), `torque_on_me` (N.m), and `signed_distance` (meters; negative means penetration). Vectors use world XYZ and act on `my_geom`. Torque is the contact's rolling/torsional torque about the contact point; it excludes the moment of the force about the body's COM. Contacts are from the most recent physics solve, or the settled initial solve on the first call. Self-contacts have one record for each of your participating geoms. A platform contact may be on its top or sides.

Each `contact_impulses` record contains `my_geom`, `other_kind`, `other_geom`, `impulse_on_me` (N.s), `torque_impulse_on_me` (N.m.s), and `position` (the force-magnitude-weighted mean contact position, or `None` if all force magnitudes were zero). These integrate force and contact torque over the preceding physics steps, including contacts that ended before the current call. Their signs and frames match `contacts`. The list is empty on the first call and resets every controller interval.

Minimal controller showing the interface and bounded return format. Neutral commands demonstrate syntax only, not navigation or combat behavior.

### controller_example

```python
def policy_step(obs) -> dict[str, float]:
    # Read the actuator names supplied by the observation.
    actuator_names = obs['my_actuator_velocity'].keys()

    # Neutral commands demonstrate the interface only.
    # Define each actuator's command according to your design and strategy.
    commands = {name: 0.0 for name in actuator_names}

    # Return a bounded float for every actuator.
    return {
        name: max(-1.0, min(1.0, float(command)))
        for name, command in commands.items()
    }
```

## DESIGN-SPACE EXPLORATION

### PERFORMANCE-FIRST DESIGN MANDATE

Optimize for the highest possible tournament win rate, not for simplicity, familiarity, ease of explanation, or conservative implementation. Qualification is a hard constraint, not the objective. A merely reliable bot that draws or loses to stronger designs is a failed design.

Take calculated engineering risks when they offer a meaningful competitive advantage. Do not choose a conventional primitive-based morphology merely because it is easier to implement or reason about. Do not treat implementation complexity, unfamiliarity, or first-attempt uncertainty as reasons to reject a stronger design.

Seriously consider custom inline SDF geometry whenever precise contact geometry could improve engagement, retention, lifting, stability, defense, locomotion, or resistance to being pushed. Primitive geoms remain acceptable only when they are genuinely the best physical solution, not when they are simply the safest or easiest option.

Design every surface intentionally. Consider whether custom profiles, concavities, tapered leading edges, asymmetric structures, interlocking contact surfaces, compliant articulation, unusual locomotion, or multifunctional mechanisms could outperform simple boxes and cylinders. Geometry should encode as much useful behavior as possible into the hardware.

Be aggressive in competitive ambition but disciplined in engineering. "Go big" does not mean using arbitrary complexity, maximum actuator gear, or unstable mechanisms. It means pursuing the strongest physically justified design, accepting manageable implementation risk, and spending complexity only where it increases expected match performance.

Before selecting the final design, ask:

- Is this the strongest morphology I can construct under the rules, or merely the easiest?
- Am I avoiding a custom shape or mechanism because it is genuinely inferior, or because it requires more careful engineering?
- What design would I choose if losing the tournament had an extreme cost?
- Does every major geometric feature improve attack, movement, defense, or recovery?
- Can the hardware itself simplify the controller or force the opponent into unfavorable contact?
- What likely opponent archetypes defeat this design, and what physical changes would remove those weaknesses?

Prefer a bold, purpose-built design with a clear path to winning over a generic design with a clear path only to surviving.

A single submission limits the number of robot designs you return, not the breadth of the design space you should explore.

Explore diverse candidate morphologies before committing to an architecture. Seek substantial differences in body organization, locomotion, support geometry, articulation, and how forces are applied to the opponent. Consider how different arrangements of physical parts could produce effective behavior, and how the same structures might contribute to movement, attack, and defense.

Assess promising candidates using the supplied physics and observations. Identify their potential advantages, dominant failure modes, and quantitative feasibility. Consider mass distribution, actuator forces, contact mechanics, stability, movement from arbitrary spawn positions, and the inactivity rule.

Compare the candidates by expected tournament win rate. Account for uncertainty, but do not let uncertainty dominate the decision: accept calculated risk when the potential competitive advantage is substantial. Select the design with the strongest physically justified path to defeating diverse opponents, not the design with the lowest implementation risk. Let this comparison determine which morphology you develop in detail.

Choose primitives, SDFs, or a combination solely according to which geometry best supports the selected design.

All examples in the syntax and observation references demonstrate implementation syntax. They do not recommend a particular morphology or control architecture.

## CONTROLLER-SPACE EXPLORATION

Explore the controller design space as seriously as the morphology space. Before committing, compare materially different ways to turn the full observation stream, observation history, and actuator feedback into winning behavior. Consider estimation, prediction, planning, adaptation, recovery, opponent behavior, and coordinated use of every performance-relevant mechanism, without treating this list as exhaustive. Select by expected tournament win rate, not ease of implementation. Do not reject stateful, sophisticated, or calculated-risk control when it offers a credible advantage. Stress-test the final controller against arbitrary spawns, diverse opponents, edge pressure, tipping, stalling, and long matches.

Use your full design + engineering reasoning budget.

Check the controller's behavior against diverse opponents, edge pressure, tipping, stalling, and long matches. Include concrete scenarios, such as: "If the robot starts at (-3, 2), facing away from its opponent, does the controller turn toward the opponent or drive off the edge?"
\end{lstlisting}

\subsection{Observation Schema}

The controller receives a Python dictionary with the following fields (excerpted verbatim from the prompt's \texttt{obs\_schema} section; nested record formats are described in the prompt below).

\begin{lstlisting}[language=,breaklines=true,basicstyle=\ttfamily\scriptsize,columns=fullflexible,keepspaces=true]
fields:
  # === POSITION ===
  my_pos: "np.ndarray[3] - entire robot's center of mass [x, y, z] in world coordinates (meters); XY is the horizontal plane, +Z is up"
  opponent_pos: "np.ndarray[3] - opponent's center of mass, using the same convention as my_pos"

  # === ORIENTATION (scalars, radians) ===
  my_yaw: "float - root-body heading in radians, from -pi to pi (0 = world +X, pi/2 = world +Y)"
  my_pitch: "float - root-body pitch in radians, from -pi/2 to pi/2 (0 = level, positive = nose down)"
  my_roll: "float - root-body roll in radians, from -pi to pi (0 = level, positive = left side up, right side down)"
  opponent_yaw: "float - opponent's root-body heading, using the same convention as my_yaw"
  opponent_pitch: "float - opponent's root-body pitch, using the same convention as my_pitch"
  opponent_roll: "float - opponent's root-body roll, using the same convention as my_roll"

  # === VELOCITY ===
  my_velocity: "np.ndarray[3] - velocity of the root-body origin [vx, vy, vz] in world coordinates (m/s)"
  my_angular_velocity: "np.ndarray[3] - angular velocity [wx, wy, wz] about the root body's local X, Y, Z axes (rad/s), positive by the right-hand rule"
  opponent_velocity: "np.ndarray[3] - velocity of the opponent's root-body origin, using the same convention as my_velocity"
  opponent_angular_velocity: "np.ndarray[3] - angular velocity about the opponent's root-body local axes, using the same convention as my_angular_velocity"
  my_actuator_velocity: |
    dict[str, float] - maps each motor's XML name to the velocity defined below for its transmission type:
    - Hinge: angular velocity about the joint axis (rad/s), positive by the right-hand rule.
    - Slide: linear velocity along the joint axis (m/s), positive in the axis direction.
    - Spatial tendon: length change per second (m/s), positive when lengthening.
    - Fixed tendon: sum(coef x joint velocity), using signed joint velocities and coefficients.
    - Ball: joint angular velocity about the motor's torque axis (rad/s), positive in the torque direction produced by a positive motor command. Gear magnitude does not scale this value; zero gear returns 0.
    - Slider-crank: slider-displacement rate in m/s, before gear scaling (0 for zero gear).
    - Relative-site: MuJoCo's scalar actuator velocity, including all gear components. It combines the selected relative translation and rotation rates; it is not an individual joint velocity.
  opponent_actuator_velocity: "dict[str, float] - opponent motor velocities, keyed by their XML names, using the same convention as my_actuator_velocity"

  # === DISTANCES ===
  distance_to_opponent: "float - Euclidean distance in XYZ between the robots' centers of mass, my_pos and opponent_pos (meters)"
  my_edge_distance: "float - signed XY distance from your robot's center of mass to the platform edge (meters); positive inside the platform boundary, zero on it, negative outside"
  opponent_edge_distance: "float - signed XY distance from the opponent's center of mass to the platform edge, using the same convention as my_edge_distance"

  # === CONTACT ===
  opponent_contact: "bool - True if any of your robot's geoms contact an opponent geom"
  opponent_contact_force: "float - sum of force magnitudes across all contacts with the opponent (N); 0 if no contact"
  ground_contact: "bool - True if any of your robot's geoms contact the platform, including its sides"

  # === STABILITY ===
  is_tipping: "float - angle in radians between the root body's +Z axis and world +Z, divided by pi/2 and capped at 1 (0 = upright, 1 = tilted 90 deg or more)"

  # === INACTIVITY ===
  my_inactivity_timer: |
    float - Duration in seconds of the longest continuous period ending now
    during which every pair of recorded center-of-mass positions is less
    than 0.5 m apart in XYZ. Starts at 0 when the round begins.
    Inactivity loss triggers at 10 seconds.
  opponent_inactivity_timer: "float - measured as described for my_inactivity_timer, but for the opponent robot. Always 0 for the qualification block."

  # === TIME ===
  t: "int - completed control steps in this match; starts at 0 on the first policy_step call of each match and increases by 1 per call. Elapsed simulated time = t x 0.01 seconds"
  max_t: |
    int - control-step limit for this match: 2000 for qualification (20 seconds), 30000 for the tournament (300 seconds).
    On every policy_step call, 0 <= t < max_t.
    Scheduled time remaining = (max_t - t) x 0.01 seconds.

  # === ROBOT PROPERTIES (static per match) ===
  my_bounding_radius: "float - approximate robot radius (meters), computed once at initialization as max(XY distance from COM to each geom's center + that geom's bounding-sphere radius). Does not update as the robot moves or articulates."
  opponent_bounding_radius: "float - computed as described for my_bounding_radius, but for the opponent robot"
  ring_radius: "float - arena radius in meters (distance from center to edge)"

  # === SPATIAL GRIDS (2D top-down maps, rebuilt every timestep) ===
  arena_grid: "np.ndarray[N,N] int8 - occupancy map of circles centered on robot COMs using the static bounding radii, not actual geometry. -1=outside ring, 0=free, 1=opponent, 2=self; self overwrites opponent where circles overlap. center cell = ring center. world_x = (col - N//2) * 0.25, world_y = (N//2 - row) * 0.25"
  arena_mass_grid: "np.ndarray[N,N] float32 - uses the same circular masks as arena_grid. -1=outside, 0=free; each occupied cell repeats that robot's entire mass (kg), not mass within that cell. Self overwrites opponent on overlap."
  edge_distance_grid: "np.ndarray[N,N] float32 - signed distance to ring edge per cell. Positive=inside, negative=outside. Static per match."

  # === GAME CONTEXT ===
  game: "dict - reserved for game-specific context; empty ({}) in every match. Your controller may ignore it."

  # === HISTORY (rolling queues, most recent last) ===
  obs_history: "list[dict] - last N observations (N=max_obs_lookback, default 20). Each dict has: my_pos, opponent_pos, my_yaw, my_velocity, opponent_velocity, distance_to_opponent, my_edge_distance, opponent_edge_distance, t, my_actuator_velocity, game. Empty at t=0."
  action_history: "list[dict] - last N actions YOU took (N=max_action_lookback, default 20). Each dict maps actuator_name -> float value. Commands are recorded before motor-specific ctrl_range limits are applied. Empty at t=0."


  # === DETAILED PHYSICAL STATE ===
  control_dt: "float - simulated seconds per controller call (0.01)"
  elapsed_time: "float - elapsed match time in simulated seconds, equal to t x control_dt"
  time_remaining: "float - scheduled time remaining in simulated seconds, equal to (max_t - t) x control_dt"
  platform: |
    dict with the following keys; all distances are in meters:
    - shape: str, "cylinder".
    - center: np.ndarray[3], center of the platform's top surface in world XYZ coordinates.
    - radius: float, platform radius.
    - half_extents: np.ndarray[2], [radius, radius] along world X and Y.
    - top_z: float, platform-top height in world coordinates; equals center[2].
    - floor_z: float, lower-floor height in world coordinates.
  my_mass: "float - total simulated robot mass, including motor mass (kg)"
  opponent_mass: "float - opponent's total simulated mass (kg)"
  my_com_velocity: "np.ndarray[3] - velocity of your entire robot's center of mass in world XYZ (m/s)"
  opponent_com_velocity: "np.ndarray[3] - opponent's center-of-mass velocity in world XYZ (m/s)"
  my_robot: "dict - named geometry, bodies, joints, tendons, motors, and sites, with definitions and current state as specified below"
  opponent_robot: "dict - the same physical information for the opponent; names belong to its XML"
  opponent_surface_distance: "float - minimum separation between the robots' collision shapes (meters), computed numerically; 0 when intersecting"
  opponent_proximity: "list[dict] - closest geom pairs with separation <= proximity_cutoff; records specified below"
  proximity_cutoff: "float - maximum separation included in opponent_proximity (2 meters)"
  proximity_limit: "int - maximum number of geom pairs returned in opponent_proximity (32)"
  proximity_truncated: "bool - True if more than proximity_limit geom pairs were within proximity_cutoff"
  contacts: "list[dict] - current solver contacts involving your robot, with positions and force/torque vectors as specified below"
  contact_impulses: "list[dict] - impulses accumulated over all physics steps since the previous controller call, grouped by geom pair"
  contact_interval: "float - simulated duration covered by contact_impulses (seconds); 0 on the first call, normally control_dt thereafter"

controller_environment: |
  Available libraries (pre-loaded, no other imports allowed):
    import math         # standard math functions (atan2, sqrt, pi, sin, cos, etc.)
    import numpy as np  # also available as bare 'np' (pre-injected into namespace)
    import typing       # type hints only
    import random       # standard PRNG (seed it yourself if you need determinism)
    import collections  # deque, defaultdict, Counter
  All Python builtins are available (`range`, `len`, `min`, `max`, `abs`, `dict`, `list`, etc.), except `open`, `exec`, `eval`, `compile`, `breakpoint`, `exit`, `quit`, `input`, `globals`, `locals`, `vars`, `getattr`, `setattr`, and `delattr`, which are forbidden. File and process access, including NumPy's save/load family, is forbidden.
  Forbidden: os, subprocess, socket, open(), exec(), eval(), pickle, threading
\end{lstlisting}

\end{document}